\documentclass[10pt,twocolumn,letterpaper]{article}

\usepackage[pagenumbers]{cvpr} 
\usepackage{amsmath}
\usepackage{algorithm}
\usepackage{algpseudocode}
\usepackage{xcolor}
\DeclareMathOperator*{\argmax}{argmax}
\DeclareMathOperator*{\argmin}{argmin}
\DeclareMathOperator*{\argsort}{argsort}
\DeclareMathOperator*{\argwhere}{argwhere}

\definecolor{cvprblue}{rgb}{0.21,0.49,0.74}
\usepackage[pagebackref,breaklinks,colorlinks,citecolor=cvprblue]{hyperref}

\title{Acquisition of Spatially-Varying Reflectance and Surface Normals via Polarized Reflectance Fields}

\author{Jing Yang$^1$, Pratusha Bhuvana Prasad$^1$, Qing Zhang$^2$, Yajie Zhao$^1$ \\
$^1$University of Southern California, $^2$Sony R\&D Center US \\
{\tt\small \{jyang, bprasad, zhao\}@ict.usc.edu}
}

\begin{document}
\maketitle
\setlength{\textfloatsep}{0.8pt}
\setlength{\abovedisplayskip}{0.5pt} 
\setlength{\belowdisplayskip}{0pt} 
\begin{abstract}
\vspace{-10pt}
Accurately measuring the geometry and spatially-varying reflectance of real-world objects is a complex task due to their intricate shapes formed by concave features, hollow engravings and diverse surfaces, resulting in inter-reflection and occlusion when photographed. Moreover, issues like lens flare and overexposure can arise from interference from secondary reflections and limitations of hardware even in professional studios. In this paper, we propose a novel approach using polarized reflectance field capture and a comprehensive statistical analysis algorithm to obtain highly accurate surface normals (within 0.1mm/px) and spatially-varying reflectance data, including albedo, specular separation, roughness, and anisotropy parameters for realistic rendering and analysis. Our algorithm removes image artifacts via analytical modeling and further employs both an initial step and an optimization step computed on the whole image collection to further enhance the precision of per-pixel surface reflectance and normal measurement. We showcase the captured shapes and reflectance of diverse objects with a wide material range, spanning from highly diffuse to highly glossy — a challenge unaddressed by prior techniques. Our approach enhances downstream applications by offering precise measurements for realistic rendering and provides a valuable training dataset for emerging research in inverse rendering. We will release the polarized reflectance fields of several captured objects with this work. 
\end{abstract}    
\vspace{-20pt}
\section{Introduction}
\label{sec:intro}

Creating realistic renderings of real-world objects is a complex task with diverse applications, including online shopping, game design, VR/AR telepresence, and visual effects. It requires precise modeling and measurement of an object's 3D geometry and reflectance properties. Recent advancements in neural renderings, such as NeRF \cite{mildenhall2020nerf} and Gaussian Splatting \cite{kerbl3Dgaussians}, offer superior realism through implicit representation but are limited to fixed scenes with fixed illumination. Ongoing research~\cite{srinivasan2021nerv, boss2021nerd, boss2021neural, yang2022light} explores relighting and inverse-rendering in neural fields. This research requires an understanding of real-world object materials, necessitating a database of material measurements. However, accurately estimating 3D geometry and reflectance properties, encompassing diffuse and specular aspects, poses a significant challenge due to the complex interplay of lighting, geometry, and spatially varying reflectance~\cite{barron2014shape}. Pioneer techniques~\cite{goldman2009shape,woodham1979photometric,basri2007photometric} combine multi-view 3D reconstruction and photography under diverse illumination to measure the geometry and spatially-varying Bidirectional Reflectance Distribution Function (SVBRDF) of real-world objects. Once measured, the models become renderable from any perspective, enabling a faithful representation of digital models in virtual environments.

Acquiring an object's reflectance properties involves measuring its SVBRDF, which requires observations from continuously changing view and lighting angles, resulting in large amounts of captured texture data. Previous works \cite{dana1999reflectance, rainer2020unified, weinmann2014material, fan2023neural} fall into this category, capturing 4D texture databases as Bidirectional Texture Functions (BTFs) and employing data-driven image-based re-rendering. These methods interpolate between sampled BRDF values from the captured 4D textures, offering accuracy but requiring substantial storage and custom interpolation functions. Moreover, they often focus on small patches, neglecting surface geometry \cite{rainer2020unified}, resulting in incompatibility with modern rendering engines. A recent discrete and sparse pattern-based approach~\cite{garces2023towards} adopts a parametric SVBRDF, achieving similar rendering quality for fabric materials compared to BTFs with a more intuitive representation. However, they optimize reflectance via SSIM loss, which may suffer from high bias in local minima.

Most material reflectance capture methods employ analytical BRDF models that rely on a few parameters, enabling sparse observations for parameter estimation and seamless integration into modern rendering pipelines for realism. Pioneering practical techniques by \cite{ma2007rapid, ghosh2011multiview} utilize programmable and polarized LEDs with multi-view DSLR cameras, efficiently separating albedo and specular components and obtaining high-fidelity surface normals via polarization and gradient illuminations. \cite{ghosh2009estimating} introduces second-order spherical gradient illumination for capturing specular roughness and anisotropy via a few captures. However, it's limited to a single viewpoint due to linearly polarized illumination that requires precise tuning for albedo-specular separation. \cite{ghosh2010circularly} further employs circularly polarized illumination to address view dependency, but fails to separate diffuse and specular reflectance or recover surface normals. In contrast, \cite{tunwattanapong2013acquiring} proposes a comprehensive technique for measuring geometry and spatially-varying reflectance under continuous spherical harmonic illumination without a polarizer. While suitable for many objects, it fails with complex, non-convex objects with interreflection and occlusion. Real-world objects with varying materials and imperfect lighting and camera conditions result in artifacts like interreflection shadows, over-exposure, and lens glare, limiting the applicability of current capture methods.

In this paper, we present a novel, practical, and precise approach for acquiring spatially-varying reflectance and object geometry. We leverage a polarized reflectance field to densely capture objects from diverse lighting directions through three steps: 1) Data Preprocessing. We analyze, model, and reduce noise arising from uncontrollable factors like overexposure, inter-reflection, and lens flare. 2) Initialization. Utilizing the preprocessed imagery, we solve for initial albedo and specular separation under gradient illumination assumptions. 3) Optimization. This stage involves optimizing surface normals, anisotropy, and roughness while updating albedo and specular maps. Our experiments showcase significantly improved capture quality and accuracy. 
In summary, our contributions include:
\begin{enumerate}
\item A unique setup for capturing the polarized reflectance fields of an object.
\item A comprehensive solution for accurately measuring the geometry (surface normal) and spatially-varying reflectance of real-world objects, encompassing albedo, specular, roughness, and anisotropy parameters.
\end{enumerate}
\vspace{-5pt}
\section{Related Works}
\label{sec:related works}

\subsection{Analytic Reflection Models}
Ward~\cite{ward1992measuring} and the simplified Torrance-Sparrow~\cite{solomon1996extracting} are widely used BRDF models in tasks to acquire reflectance field~\cite{nam2018practical, debevec2000acquiring}. While the simplified Torrance-Sparrow addresses isotropic reflection only, the Ward model, a simplified version of the Cook-Torrance model~\cite{cook1982reflectance}, is physically valid for both isotropic and anisotropic reflections. Additionally, BRDFs have been applied in recent physically based differentiable rendering techniques, often with certain approximations such as Blinn-Phong~\cite{li2018differentiable}, isotropic specular reflection~\cite{cai2022physics}, cosine-weighted BRDF~\cite{zhang2019differential}.

SVBRDFs encompass 2D maps of surface properties such as texture and roughness. Most studies focus on SVBRDF acquisition of planar surfaces~\cite{hui2017reflectance, li2017modeling, ye2018single, deschaintre2018single, gao2019deep, ma2023opensvbrdf}. For non-planar objects, \cite{li2018learning} predicted both shape and SVBRDF from a single image, but with limited photo-realism.
Utilizing polarization cues under flash illumination, \cite{deschaintre2021deep} achieved higher quality in specular effect but suffers from inaccurate diffuse albedo due to baked-in specular highlights. \cite{hwang2022sparse} captured the polarimetric SVBRDF, including the 3D Mueller matrix, yet lacked anisotropic specular effects. A recent work~\cite{garces2023towards} captured both anisotropic reflectance at the microscopic level and employed an image translation network to propagate BRDF from micro to meso, successfully fitting specular reflectance without diffuse lobe influence. However, they did not decouple and explicitly optimize the specular parameters.

\vspace{-2pt}
\subsection{ML-based BTF capture}
Recently, the bidirectional texture function (BTF) has been introduced to model finer reflection including mesoscopic effects such as subsurface scattering, interreflection, and self-occlusion across the surface~\cite{dana1999reflectance}. Recent advancements utilize neural representations trained to replicate observations, as BTF lacks an analytical form.
\cite{weinmann2014material} synthesized BTF under different views and illuminations and trained an SVM classifier on the synthesized dataset to classify real-world materials based on a low-dimensional feature descriptor. ~\cite{rainer2020unified} trained an autoencoder that simulated discrete angular integration of the product of the reflectance signal with angular filters by projecting 4D light direction and RGB into a weighted matrix and then encoding them into a latent vector. The decoder outputs RGB from the latent vector concatenated with query directions. Upon this, \cite{kuznetsov2021neumip} employed a neural texture pyramid instead of the encoder to represent multi-scale BTFs, achieving smaller storage but more levels of detail, i.e., accurate parallax and self-shadowing. \cite{kuznetsov2022rendering} added surface curvature into BTF input and outputs opacity alongside RGB color. It also allowed for UV coordinate offset to handle silhouette and parallax effects for near-grazing viewing directions. \cite{fan2023neural} presented a biplane representation of BTF, including spatial and half feature maps, and employed a small universal MLP for radiance decoding, achieving a faster evaluation compared to the method \cite{kuznetsov2022rendering}. Nonetheless, current BTF representations lack the flexibility and generality of SVBRDF. Aiming for the standard industrial rendering pipeline, we adopt SVBRDFs in this work. Although mesoscopic effects are beyond our focus, our approach does not contradict with any potential BTF extension.

\subsection{Gradient Illumination for BRDF}
We exclusively concentrate on polarized illumination, as it overtakes non-polarized approaches in acquiring the specular reflectance properties.
\cite{ma2007rapid} designed the methodology to separate diffuse and specular components and obtain corresponding normals from polarized 1st-order spherical gradient illumination patterns. While it assumes that the object is isotropic and has a small specular lobe throughout, \cite{ghosh2009estimating} made a weaker specular BRDF assumption, only symmetry about the mean direction, and derived computation of roughness and anisotropy from the 2nd-order spherical gradient illumination. \cite{ghosh2010circularly} separated and inferred the specular roughness from circularly polarized illumination using the Stokes vector parameters. \cite{ghosh2011multiview} degraded the linear polarized pattern of \cite{ma2007rapid} to two latitudinal and longitudinal patterns, allowing diffuse-specular separation for multiview stereo captures. The obtained photometric normals are then used to constrain further stereo reconstruction. \cite{tunwattanapong2013acquiring} adopted up to 5th-order continuous spherical harmonic illumination to obtain diffuse, specular roughness and anisotropy. Instead of optimization, it used a table of a range of roughness and anisotropy to integrate the 5th-order spherical harmonics and found the best-matched specular parameters. With color spherical gradient illuminations and linear polarizers placed on cameras, \cite{fyffe2015single} could acquire diffuse and specular albedo and normals simultaneously at a single shot with a Phong model. \cite{guo2019relightables} addressed the unrealistic double shading issue in this single-shot approach using two different color gradient illuminations. \cite{legendre2018efficient} argued the complexity of using both color lights and cameras, adopting monochrome cameras that can still hallucinate parallel- and cross-polarized images under unpolarized illuminations. \cite{kampouris2018diffuse} adopted binary gradient illumination, which requires fewer photos than spherical gradient illumination. As discussed in Section~\ref{sec:intro}, our capture is based on the linear polarized spherical gradient illumination~\cite{ma2007rapid}.

\vspace{-2pt}
\section{Preliminary}

\paragraph{Fresnel Equations}
The Fresnel equations describe how the light behaves when it encounters the boundary between different optical media, involving two primary reflection components: specular and diffuse reflection. Specular reflection occurs unscattered, producing distinct reflections at any interface. In contrast, diffuse reflection results from both surface and subsurface scattering, causing light to scatter in various directions. The Fresnel equation specifies that specular reflection retains the polarization state of the incident light, while diffuse reflection remains unpolarized, regardless of the incident light's polarization characteristics \cite{ma2007rapid}.
Therefore, diffuse and specular reflection can be separated with different states of polarization, which are determined by incident light conditions. 

\vspace{-10pt}
\paragraph{Linear Polarization and Malus' Law}
Theoretically, placing linear polarizers and analyzers in front of light sources and observers with different orientations can effectively separate diffuse and specular reflections. When the polarizer and analyzer are set perpendicular to each other, only the diffuse reflection becomes visible, and the intensity of polarized light is governed by Malus's Law:
\begin{equation}
I= I_0 \cos^2\theta
\end{equation}
where $\theta$ is the angle between the polarizer's axis and the analyzer's axis. Given that the average of $\cos^2\theta$ is $\frac{1}{2}$, under identical lighting conditions, the radiant intensity of diffuse reflection $I_d$ and specular reflection $I_s$ can be measured by: 
\begin{equation}
I_d= 2I_\perp,\quad I_s= 2I_\parallel - 2I_\perp
\end{equation}
where $I_\perp$ and  $I_\parallel$ are the observations under cross and parallel-polarized lighting respectively; proof can be found in supplementary material via Mueller calculus. 

\vspace{-10pt}
\paragraph{Task Statement} 
Materials are represented using spatially-varying BRDF, which explains light scattering on a material's surface in various directions.  Our goal is to precisely measure the following reflectance attributes with controllable polarized lighting: diffuse and specular albedo $\rho_d$, $\rho_s$, diffuse and specular normal $n_d$, $n_s$, specular variance $\sigma$, anisotropy $\varsigma$, and roughness $\gamma$. We also measure diffuse and specular visibility $\nu_d$, $\nu_s$, inter-reflection $\varsigma_d$, $\varsigma_s$, and occlusion $\tau_d$, $\tau_s$. 

\vspace{-10pt}
\paragraph{Data Format} Table \ref{tab:symbols} summarizes the symbols and formats of the input and output data involved in our method. $\mathbb{R}_{+}$ encompasses all non-negative real numbers, and $\mathbb{B}=\{0, 1\}$. $H$ and $W$ respectively refer to the height and width of the image. In our case, the capture is executed with 8 RED KOMODO 6K cameras at $H=6144$ and $W=3240$ at 30 FPS covering $N=346$ lighting directions.
\begin{table}[t]
    \centering
    \vspace{-15pt}
    \small
    \begin{tabular}{clcl} 
          \hline
          I/O&Name &  Symbol&Dimension \\
          \hline
 I& Camera Pose & $\mathtt{R}, \mathtt{t}, \mathtt{K}$ &$\mathbb{R}^{3 \times 3}$, $\mathbb{R}^3$, $\mathbb{R}^{3 \times 3}$ \\ 
  I&Captured Image & $I_\perp$, $I_\parallel$&$\mathbb{R}^{H \!\times\! W \!\times\! 3}$ \\ 
  I&Captured OLAT & $\Lambda_\perp$, $\Lambda_\perp$&$\mathbb{R}^{N \!\times\! H \!\times\! W \!\times\! 3}$ \\ 
  O&Visibility Map & $\nu_d$, $\nu_s$&$\mathbb{B}^{H \!\times\! W}$ \\
 O& Occlusion Map & $\tau_d$, $\tau_s$&$\mathbb{R}_{+}^{H \!\times\! W}$ \\ 
  O&Inter-reflection Map & $\varrho_d$, $\varrho_s$&$\mathbb{R}_{+}^{H \!\times\! W \!\times\! 3}$, $\mathbb{R}_{+}^{H \!\times\! W}$ \\ 
  O&Albedo Map & $\rho_d$, $\rho_s$& $\mathbb{R}_{+}^{H \!\times\! W \!\times\! 3}$, $\mathbb{R}_{+}^{H \!\times\! W}$ \\ 
  O&Normal Map & $n_d$, $n_s$& $\mathbb{R}^{H \!\times\! W \!\times\! 3}$ \\
 O& Specular-var Map& $\sigma$&$\mathbb{R}_{+}^{H \!\times\! W \times 2}$ \\ 
  O&Anisotropy Map & $\varsigma$&$\mathbb{R}_{+}^{H \!\times\! W}$\\ 
  O&Roughness Map & $\gamma$&$\mathbb{R}_{+}^{H \!\times\! W}$ \\
\end{tabular}
\vspace{-8pt}
\caption{\textbf{Symbols.} Dimensions are indicated only once if the symbols with different subscripts are within the same dimension.}
\label{tab:symbols}
\end{table}


\vspace{-5pt}
\section{Methods}
\label{sec:methods}

\vspace{-5pt}
\paragraph{Overview} The complex interplay of diffuse and specular reflections in light transport challenges accurate material capture, often resulting in imprecise measurements. To address this issue, we conduct a thorough capture process and employ statistical and optimization methods on the recorded sequences for precise material acquisition.

Our method consists of three primary steps: We start with polarized OLAT (One Light at A Time) captures, encompassing cross and parallel polarization conditions (Section \ref{sec:method-capture}). Next (Section \ref{sec:method-preprocess}), we analyze and preprocess the captured sequence to eliminate overexposure. Additionally, we define a set of constraints aimed at reducing the influence of inter-reflection, self-occlusion, and lens flare during the subsequent optimization process. Finally (Section \ref{sec:method-optimization}), we delve into the specifics of our optimization approach, dedicated to enhancing the accuracy of material properties.

\begin{figure}[t]
    \centering
    \vspace{-20pt}
    \includegraphics[width=\linewidth]{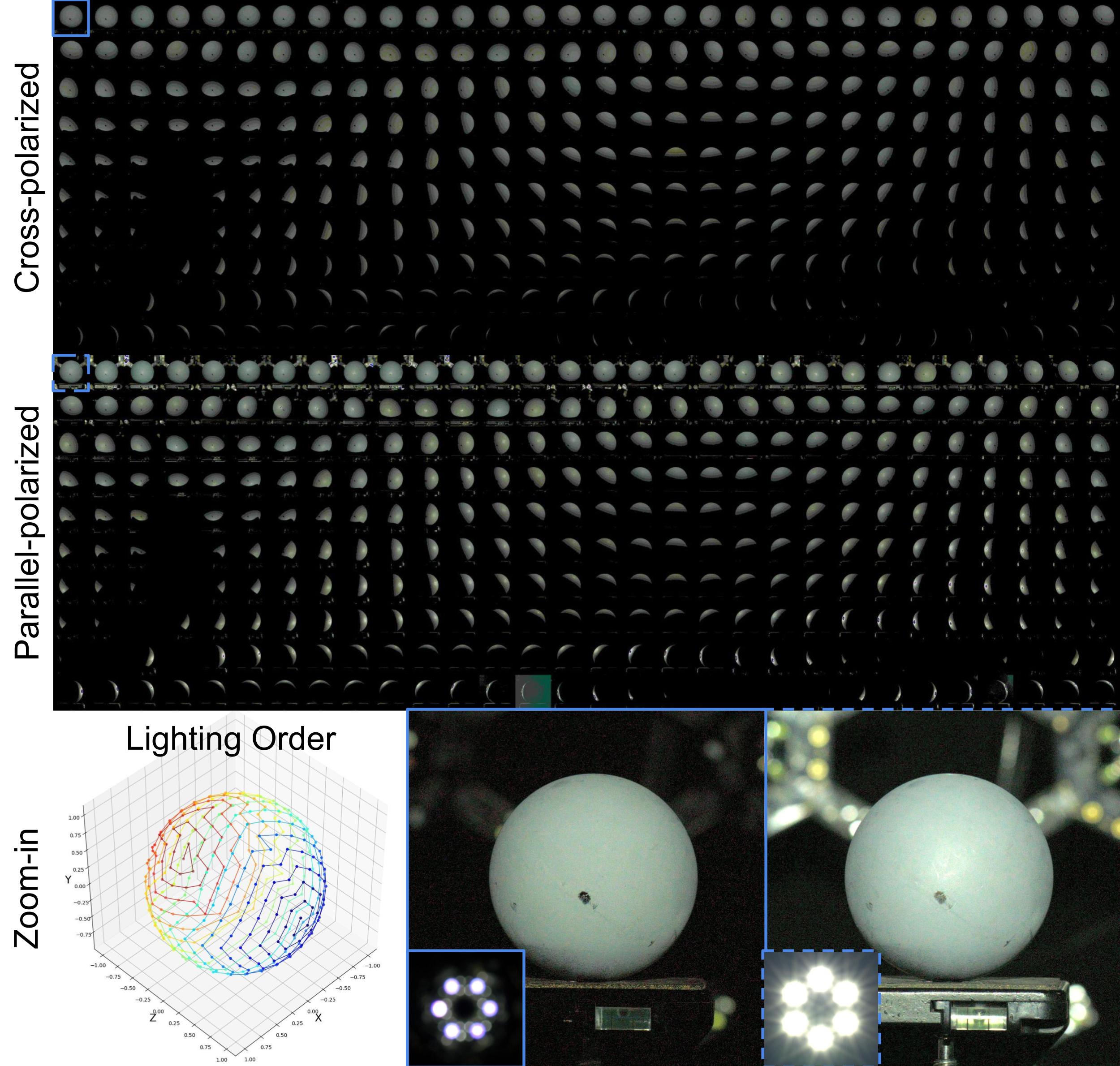}
    \vspace{-18pt}
    \caption{\textbf{Polarized OLAT.} We present an object captured under the first 300 lighting conditions. The top two rows exhibit images captured through cross-polarized OLAT and parallel-polarized OLAT. In the last row, we zoom in on details, marked by corresponding continuous/dotted boxes, with the active light board in the lower corner accordingly. Additionally, we present the lighting order, progressing from blue to red.
    }
    \vspace{3pt}
    \label{fig:polarized_olat}
\end{figure}
\vspace{-3pt}
\subsection{Polarized OLAT Capture}
\label{sec:method-capture}
\vspace{-4pt}
To separate diffuse and specular reflection, we perform the multiview captures of the same object under cross-polarized and parallel-polarized OLAT illuminations at the same intensity. The capture results in a cross-polarized sequence $\Lambda_\perp=\{I_\perp^k\}_{k=0}^{N}$ and a parallel-polarized sequence $\Lambda_\parallel=\{I_\parallel^k\}_{k=0}^{N}$ are shown in Fig.~\ref{fig:polarized_olat}, where $N$ is the length of the sequence. Therefore, the diffuse reflection sequence $\Lambda_d$ and specular reflection sequence $\Lambda_s$ can be defined as:
\vspace{5pt}
\begin{equation}
    \Lambda_d=\{2I_\perp^k\}_{k=0}^{N}, \quad \Lambda_s=\{2I_\parallel^k - 2I_\perp^k\}_{k=0}^{N}
\vspace{5pt}
\end{equation}
The transition of lighting polarization states is achieved by controlling the activation of different lights at each instance on the light board. On each light board, the white lights are arranged in a hexagonal pattern, with cross-polarizers and parallel polarizers placed alternately at the front. Throughout the capture process, each light of corresponding polarization states is activated via a 12-bit intensity code. Additionally, each OLAT sequence follows a spiral order from $+z$ to $-z$ covering all available lighting directions $\omega_i$ over the sphere, in total $N$ directions. The direction of outgoing radiance $\omega_o$ is determined by the camera pose $[\mathtt{R}, \mathtt{t}, \mathtt{K}]$ from multiview camera calibration. 

\subsection{Analysis and Preprocess}
\label{sec:method-preprocess}
While analyzing images under OLAT illumination, material observation may encompass overexposure highlights, inter-reflection, and self-occlusion. These physical phenomena can introduce inaccuracies in material measurement. In the subsequent paragraph, we visually elaborate on these effects and explain how we mitigate their impact.



\begin{figure}[t]
    \centering
    \vspace{-23pt}
    \includegraphics[width=\linewidth]{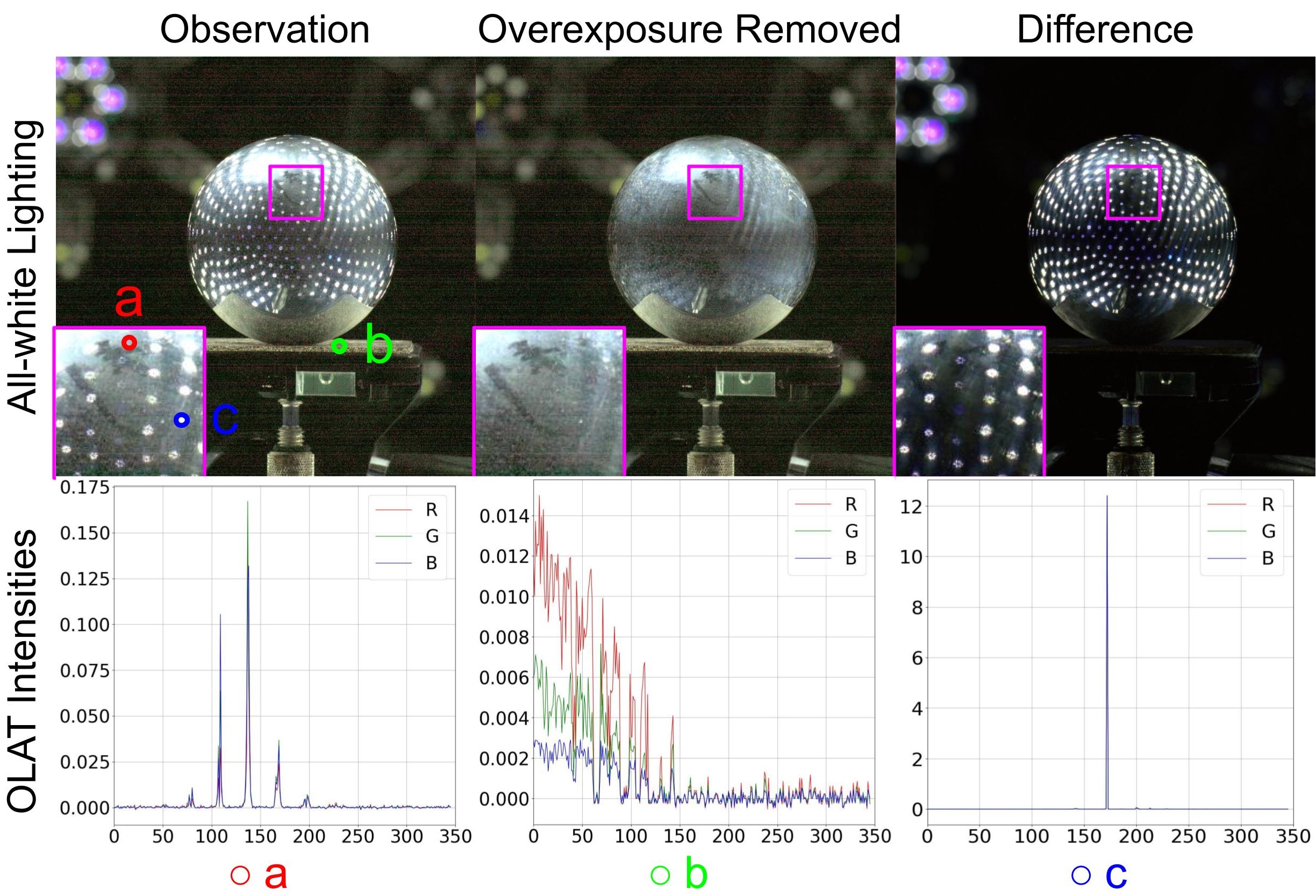}
    \vspace{-20pt}
    \caption{\textbf{Overexposure Removal.} We demonstrate the effectiveness of overexposure elimination by using a mirrorball. The observation under all-white lighting is a summation of all frames from the cross-polarized OLAT data \(\Lambda_\perp\). We select points \textcolor{red}{\textbf{a}} and \textcolor{blue}{\textbf{c}} from the zoom-in region and \textcolor{green}{\textbf{b}} from the base, which is made of relatively diffuse material. The second row presents the corresponding intensities recorded under cross-polarized OLAT lighting. The horizontal axis represents the OLAT index, while the vertical axis indicates the recorded intensities in red, green, and blue.}
    \vspace{-10pt}
    \label{fig:overexposure}
\end{figure}
\vspace{-5pt}
\begin{algorithm}[t]
\caption{Overexposure Removal}
\label{alg:overexposure}
\begin{algorithmic}
\small
\State $\text{iter} \gets 0; M \gets 2; \delta \gets \text{mean}(\Lambda)$ \;
\While{$\text{iter} < M$}
    \For{each pixel $(i,j)$}
        \State $\lambda \gets \Lambda^{(i,j)}$
        \State $\text{id}_\text{desc} \gets \argsort(\lambda)$
        \State $\Delta_\text{desc} \gets \lambda[\text{id}_\text{desc}[0:\text{end}\!-\!1]]-\lambda[\text{id}_\text{desc}[1:\text{end}]]$
        \State $\text{id}_{>\varepsilon} = \argwhere(\Delta_\text{desc} > \varepsilon)$
        \State $\lambda[\text{id}_{>\varepsilon}[0]] \gets \lambda[\text{id}_\text{desc}[1]] + \delta$
    \EndFor
    \State $\text{iter} \gets \text{iter} + 1$
\EndWhile
\vspace{5pt}
\end{algorithmic}
\end{algorithm}

\vspace{-5pt}
\paragraph{Overexposure}
Overexposure occurs when intense light interacts with a material, producing highlights that mimic the pattern of the light sources.

For a particular surface point, the intensity observation should exhibit continuity within a specific range rather than displaying an emergent highlight. Exploiting this property allows for the immediate identification of abnormal pulses indicative of overexposure. As shown in Fig.~\ref{fig:overexposure}, the intensity variations of three surface points under OLAT illumination clearly reveal the recognition of overexposure through strong pulses in the sequence.

Our approach, formulated in Algorithm \ref{alg:overexposure}, consists of sorting the signal intensities in each color channel according to the lighting order, detecting differences that exceed a predefined threshold $\varepsilon$, and replacing the values at such points with an ambient value $\delta\ll\varepsilon$. Usually, the threshold $\varepsilon$ relates to the light sources and can be easily determined during the capture. The captured data at each pixel position $(i,j)$ within the OLAT sequence $\Lambda \in \{\Lambda_d, \Lambda_s\}$ is treated as a time-domain signal $\lambda = \Lambda^{(i,j)} \in \mathbb{R}^{N\times 3}$ and this procedure can be iterated multiple times $M$ for a clean result. 

As a result, overexposure can be effectively eliminated, yielding the calibrated OLAT sequence $\hat{\Lambda} \in\{\hat{\Lambda}_d,\hat{\Lambda}_s\}$. As depicted in Fig.~\ref{fig:overexposure}, our method successfully isolates overexposure from the original capture. 

\vspace{-15pt}
\begin{figure}[t]
    \centering
    \vspace{-20pt}
    \includegraphics[width=\linewidth]{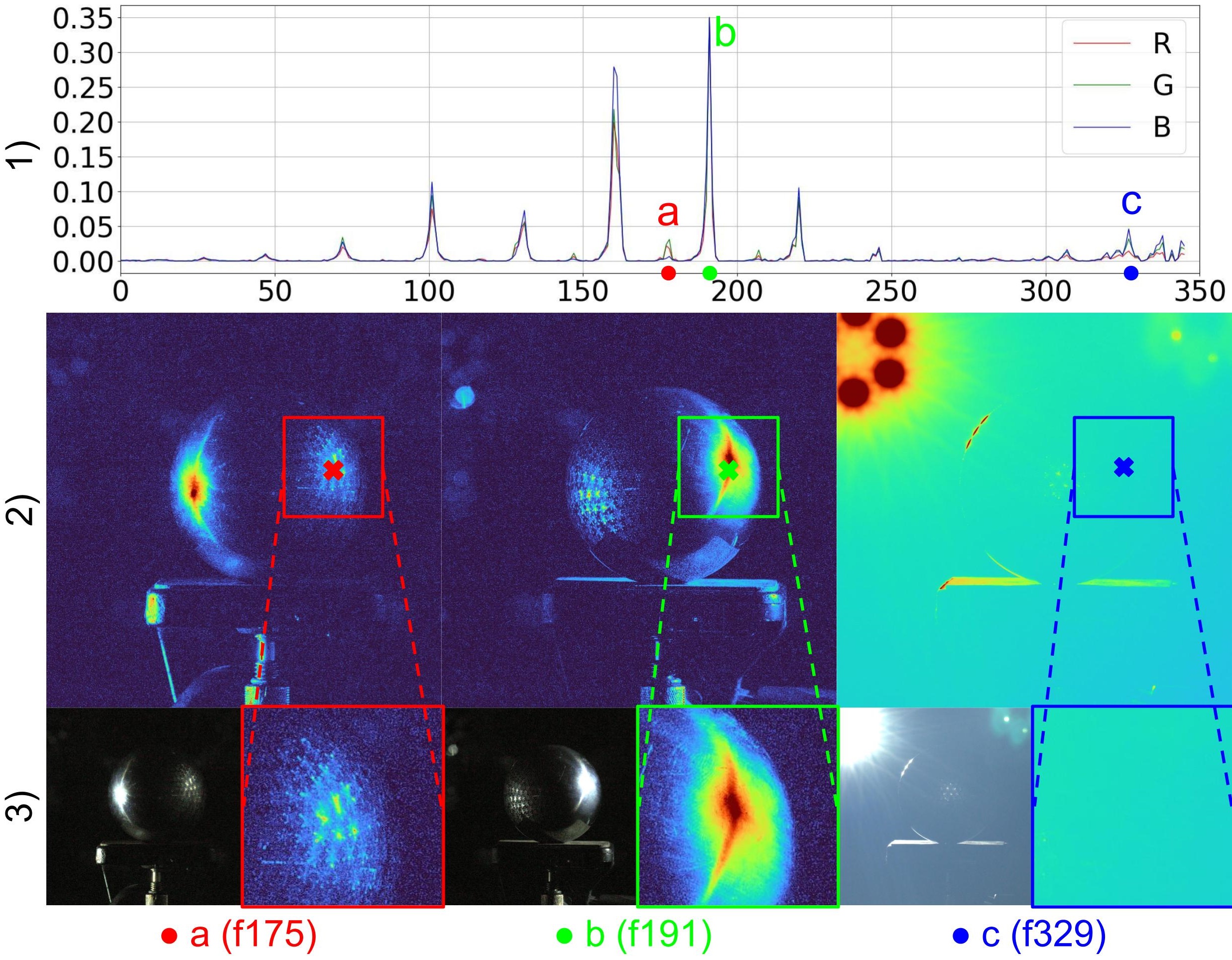}
    \vspace{-23pt}
    \caption{\textbf{Inter-reflection and Lens Flare.} In row 1), we present the captured intensity distribution of a fixed surface point, indicated with a cross ($\mathbf{\times}$) in the following example images. The intensity patterns are identified as \textcolor{red}{a}) interreflection, \textcolor{green}{b}) regular specular reflection, and \textcolor{blue}{c}) lens flare. Row 2) provides a false-color view, with a zoom-in view in the last row 3), as well as the raw capture. In the false-color view, the intensity strength is represented by the color, with stronger intensities appearing redder.}
    \vspace{3pt}
    \label{fig:interreflection}
\end{figure}

\paragraph{Inter-reflection}
Moreover, the intricate behavior of light as it bounces around often gives rise to inter-reflection, particularly in the presence of highly specular objects in the environment. This can introduce additional errors when measuring material properties. Inside the capture device, inter-reflection predominantly occurs from the reflection of light sources bouncing off the capturing layout and being captured by the camera as depicted in Fig.~\ref{fig:interreflection}. 
Furthermore, objects with concave geometries tend to exhibit a higher incidence of inter-reflection. 

When the lighting arises from the lower hemisphere $\Omega^{-}$, the object becomes unobservable due to the absence of incoming radiance. However, such a point can still be captured in the sequence caused by inter-reflection from $\omega_{ir}$. This phenomenon typically occurs on the opposite side of the direction of the active lights $\omega_i$, leading to \(\omega_{ir}\cdot\omega_i<0\). Usually, the surface point is observable when \(n\cdot\omega_i>0\).
By introducing the visibility $\nu$, the inter-reflection can be approximated via: 
\begin{equation}
\label{eqn:iterrreflection}
\begin{split}
\varrho &= \int_{\Omega^-}\nu(\omega_i, \omega_o) R(\omega_{i}, n)d\omega_{i} \\
&\approx \sum\nolimits_{k=0}^N \lceil \hat{I}^k-\zeta \rceil \cdot max(-\omega_i^k\cdot n,0)\cdot \hat{I}^k
\end{split}
\end{equation}

where $\nu(\omega_i, \omega_o)$ is the visibility of  $\omega_i$ observed via $\omega_o$, $R(\omega_{ir}, n)$ is the surface reflection function, $\zeta$ is the average ambient noise map, $\lceil\cdot\rceil$ is the ceiling operator, $n$ is the surface normal and $\hat{I}^k$ is the k-th image from the calibrated OLAT sequence $\hat{\Lambda}\in\{\hat{\Lambda}_d, \hat{\Lambda}_s\}$, where different polarized OLAT yields diffuse inter-reflection $\varrho_d$ and specular inter-reflection $\varrho_s$ accordingly. Given such observations, this inter-reflection can be mitigated by simply imposing a constraint on the incident lighting at each optimization step, specifically \(n\cdot\omega_i\geq 0\).

\begin{figure}[t]
    \centering
    \vspace{-20pt}
    \includegraphics[width=\linewidth]{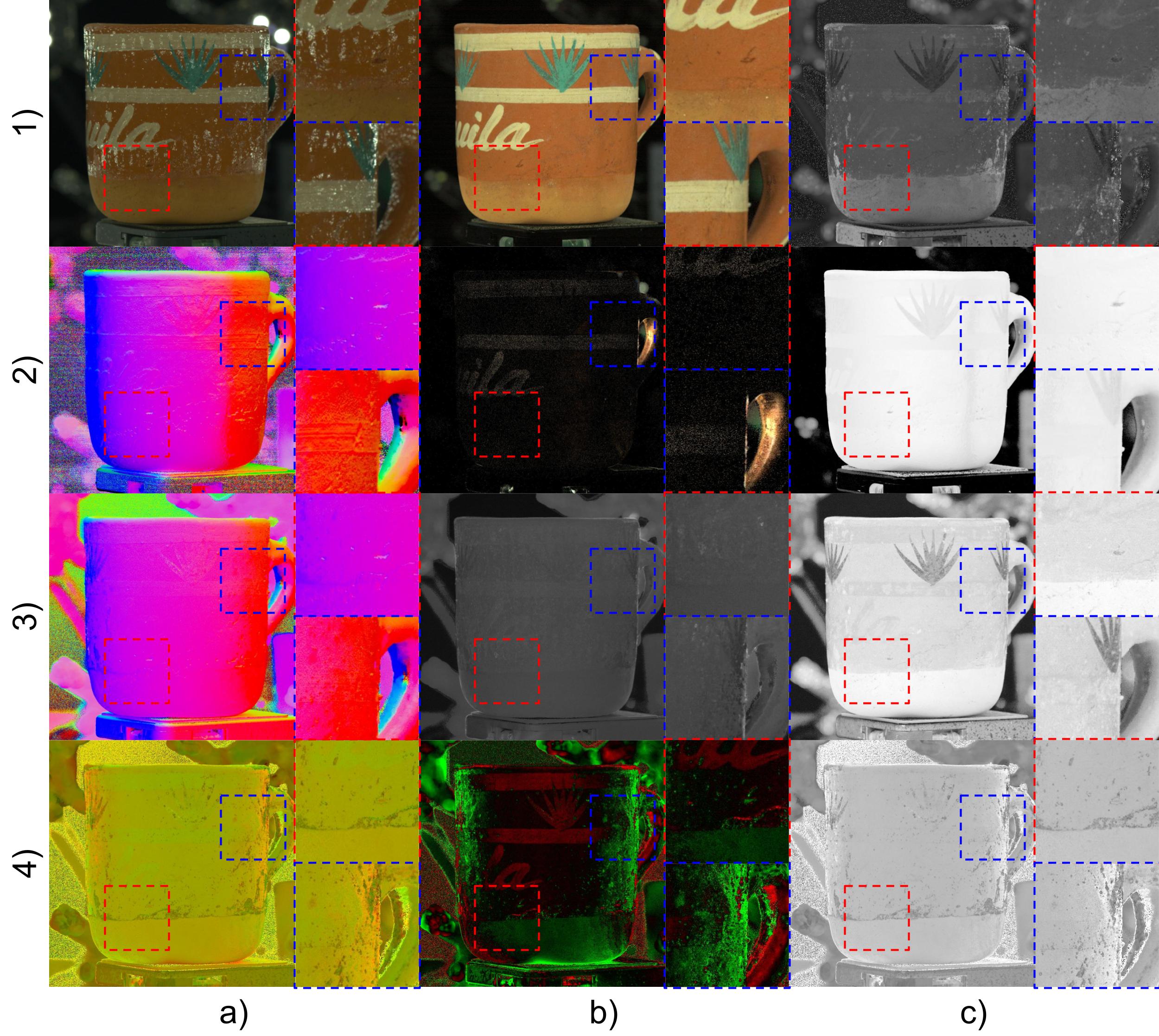}
    \vspace{-23pt}
    \caption{\textbf{Acquisition.} We present a measured mug comprising a diffuse base and a clear coat showcasing a1) original, b1) diffuse albedo $\rho_d$, c1) specular albedo $\rho_s$, a2) diffuse normal $n_d$, b2) diffuse inter-reflection $\varrho_d$ (intensity adjusted for better visualization), c2) diffuse occlusion $\tau_d$, a3) specular normal $n_s$, b3) specular inter-reflection $\varrho_s$, c3) specular occlusion $\tau_d$, a4) specular variance $\sigma$, b4) anisotropy $\varsigma$, and c4) roughness $\gamma$.}
    \vspace{2pt}
    \label{fig:decomposition}
\end{figure}

\vspace{-10pt}
\paragraph{Lens Flare}
Additionally, lens flare arises when the lighting direction aligns with the opposite side of the viewing angle $\omega_o$ as shown in Fig.~\ref{fig:interreflection}. This condition yields in \(\omega_i\cdot\omega_o<0\) and, as a result, \(n\cdot\omega_i<0\) when the surface point is observable via $\omega_o$. The effect of lens flare can therefore be reduced with the constraint \(n\cdot\omega_i\geq 0\).

\vspace{-10pt}
\paragraph{Occlusion} 
Another phenomenon that can impact material acquisition is self-occlusion, leading to the presence of shadows in the observations. The more shadows appear in the observations, the less accurate the material measurements become. The occlusion, $\tau$, is defined via:
\begin{equation}
\begin{split}
\tau&=\int_{\Omega}\nu(\omega_i, \omega_o) \cdot (\omega_i\cdot n) d\omega_i \cdot \left(\int_{\Omega} (\omega_i \cdot n) d\omega_i \right)^{-1} \\ 
&\approx \frac{4}{N}\sum\nolimits_{k=0}^{N} \lceil \hat{I}^k-\zeta \rceil \cdot max(\omega_i^k\cdot n,0)
\end{split}
\end{equation}
where $\omega_i$, $\omega_o$, $n$, $\nu(\omega_i, \omega_o)$, $\zeta$, and $\hat{I}^k$ are the same as in Equation \ref{eqn:iterrreflection}. $4/N$ is a factor to normalize the average occlusion in the hemisphere (discussed in supplementary material). The measured occlusion map can be used further to eliminate shadows from the albedo map. 

\vspace{-5pt}
\begin{figure}[t]
    \centering
    \vspace{-20pt}
    \includegraphics[width=\linewidth]{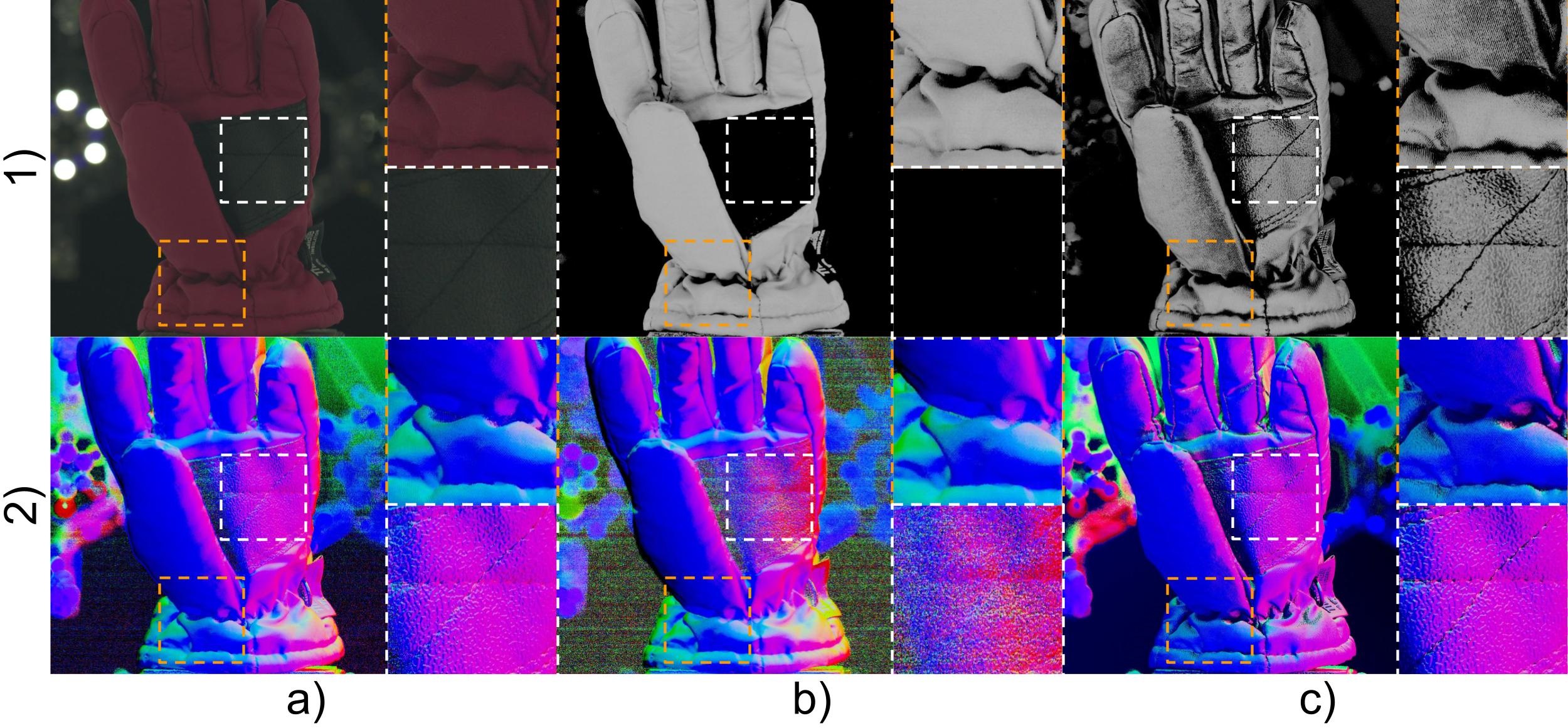}
    \vspace{-17pt}
    \caption{\textbf{Normal Fusion.} a1) original image, b1) maximum cross-correlation of diffuse normal optimization, c1) maximum cross-correlation of specular normal optimization, a2) fused normal, b2) diffuse normal, c2) specular normal.}
    \vspace{5pt}
    \label{fig:normal_fusion}
\end{figure}

\subsection{Optimization}
\label{sec:method-optimization}
Following the preprocess, our focus shifts to solving optimization problems to obtain the required material. The process begins with an \textit{initial approximation} of the solution, achieved through the synthesized gradient illumination. This is followed by successive refinements of \textit{surface normals $n$, anisotropy $\varsigma$, roughness $\gamma$, and albedo $\rho$}, each step methodically integrating constraints that emerge from our analytical evaluations. An illustration of the acquired material and intermediate outcomes is presented in Fig.~\ref{fig:decomposition}. Full derivation can be found in supplementary material.

\vspace{-10pt}
\paragraph{Initialize $\rho$ and $n$} Initial diffuse albedo $\rho_d^\text{init}$ and specular albedo $\rho_s^\text{init}$ can be easily derived from the preprocessed sequences $\hat{\Lambda}_d$ and $\hat{\Lambda}_s$ via:
\begin{equation}
\rho_d^\text{init}=\int_\Omega R_d(\omega_i, \omega_o) d\omega_i \approx \frac{4\kappa}{N} \sum\nolimits_{k=0}^N \hat{I}_d^k
\end{equation}
\begin{equation}
\rho_s^\text{init}=\int_\Omega R_s(\omega_i, \omega_o) d\omega_i \approx \frac{4\pi\kappa}{N} \sum\nolimits_{k=0}^N \hat{I}_s^k
\end{equation}
where $R_d$  and $R_s$ represent the diffuse reflection function and the specular reflection function of the material, respectively. $\kappa$ is a constant determined by light intensity and solid angle in the capture device.

Furthermore, the initial estimates for the diffuse surface normal $n_d^\text{init}$ and specular normal $n_s^\text{init}$ can be approximated from the spherical gradient illumination pattern $P_j\in \{P_x,P_y,P_z\}$ \cite{ma2007rapid}. The response under gradient lighting pattern, including negative values, $\hat{I}_{d,j} \in \{\hat{I}_{d,x}, \hat{I}_{d,y}, \hat{I}_{d,z}\}$ and $\hat{I}_{s,j} \in \{\hat{I}_{s,x}, \hat{I}_{s,y}, \hat{I}_{s,z}\}$ can be easily synthesized using the captured OLAT sequence with weighting  $w\in[-1,1]$ over the incident lighting direction. The derivation and synthesized results can be found in the supplementary material.
\begin{equation}
\hat{I}_{d,j}= \int_\Omega \!P_j(\omega_i) R_d(\omega_i, \omega_o) d\omega_i \approx \sum\nolimits_{k=0}^N w_j^k \hat{I}_d^k
\end{equation}
\begin{equation}
\hat{I}_{s,j}= \int_\Omega \!P_j(\omega_i) R_s(\omega_i, \omega_o) d\omega_i \approx \sum\nolimits_{k=0}^N w_j^k \hat{I}_s^k
\end{equation}
Moreover, with $\mathfrak{N}(\cdot)$ representing the normalization operator, the surface normal can be derived as follows:
\begin{equation}
\label{eqn:ma2007_normals}
\begin{split}
n_d^\text{init}&=\mathfrak{N}\left(\frac{3}{2\pi\rho_d} \cdot [\hat{I}_{d,x}, \hat{I}_{d,y}, \hat{I}_{d,z}]\right) \\
n_s^\text{init}&=\mathfrak{N}\big(\mathfrak{N}[\hat{I}_{s,x}, \hat{I}_{s,y}, \hat{I}_{s,z}]) + \omega_o\big) \\
\end{split}
\end{equation}



\begin{figure}[t]
    \centering
    \vspace{-20pt}
    \includegraphics[width=\linewidth]{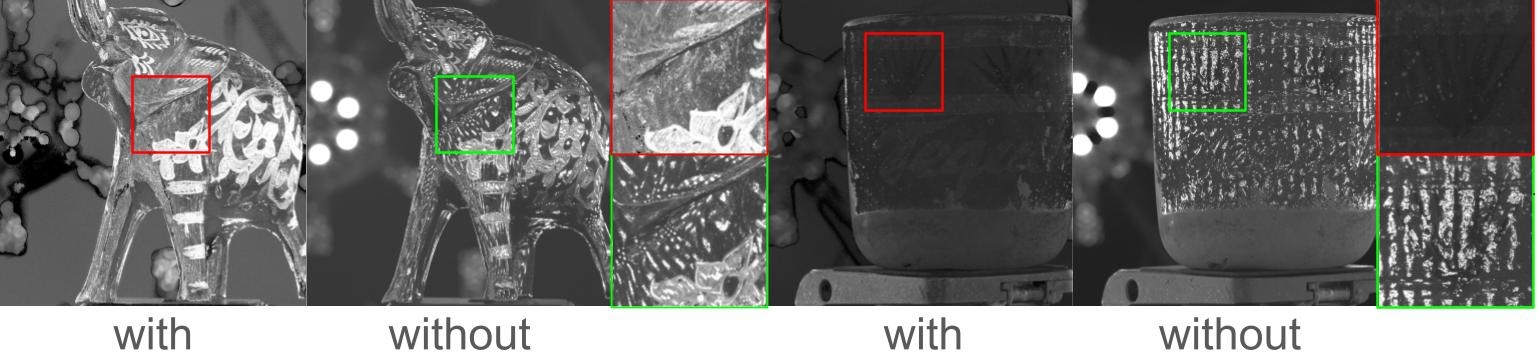}
    \vspace{-17pt}
    \caption{
    \textbf{Ablation study on overexposure removal.} The proposed overexposure removal effectively mitigates the lighting baked-in effect from the acquired specular albedo.
    }
    \label{fig:ablation_overexposure}
\end{figure}


\begin{figure}[t]
    \centering
    \vspace{-12pt}
    \includegraphics[width=\linewidth]{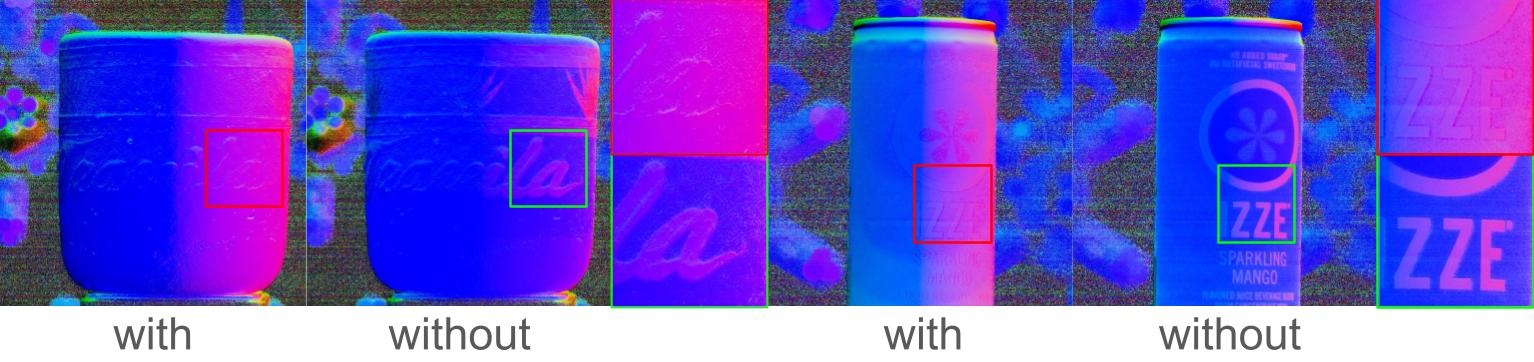}
    \vspace{-17pt}
    \caption{\textbf{Ablation study on optimization.} Without optimization, the acquired diffuse normal incorrectly reflects texture as normal. This shows much better separation after our proposed optimization step.}
    \vspace{2pt}
    \label{fig:ablation_optimization}
\end{figure}


\vspace{-10pt}
\paragraph{Refinement on $n$} 
$\hat{\Lambda}_d$ captures the diffuse characteristics of surface points under periodic illumination. The recorded data represents the area element on the object, \(dA\), observed through an aperture of area \(dA_o\), thus subtending a solid angle \(d\Omega_o\). For any arbitrary equal angle \(d\Omega\), each surface point within $\hat{\Lambda}_d$ is expected to exhibit Lambertian appearance in accordance with Lambert's law:
\begin{equation}
\label{eqn:lambert-cosine-law}
\begin{split}
L_o = \frac{L_i(\omega_i\cdot n)(\omega_o\cdot n) d\Omega dA}{d \Omega_o (\omega_o \cdot n) d A_o}=\frac{L_i(\omega_i \cdot n) d\Omega dA}{d\Omega_o dA_o}
\end{split}
\end{equation}
where $L_i$ is the incident radiance from the light source and $L_o$ denotes the observed radiance. By maintaining uniformity in $L_i$ across the reflection sequence $\hat{\Lambda}_d$, the observation of a surface point is solely influenced by the incident lighting direction $\omega_i$ and the surface normal $n$. Consequently, we can get optimized normal $\hat{n}_d$ by maximizing the overall cross-correlation with the observation $\hat{\Lambda}_d$ at each surface point through the expression: 
\begin{equation}
\begin{gathered}
\label{eqn:nd_cross_corr}
\!\hat{n}_d \!=\! \argmax_{\substack{n \\ n^\text{init}:=n_d^\text{init}}} \mathfrak{N}\left(\!\begin{bmatrix} \nu_d^0 \!\cdot\! n \!\cdot\! \omega_i^0 \\ \cdots \\ \nu_d^k \!\cdot\! n \!\cdot\! \omega_i^k \end{bmatrix}\!\right)^{\!T\!} \!\mathfrak{N}\left(\!\begin{bmatrix} \hat{I}_d^0 \\ \cdots \\ \hat{I}_d^k \end{bmatrix}\!\right) \\
\text{s.t.} \quad \lVert n \rVert=1 ,  n \cdot \omega_i^k > 0 \text{, and\,} \hat{I}_d^k \in \hat{\Lambda}_d
\end{gathered}
\end{equation}
Likewise, the refined specular normal \(\hat{n}_s\) can be attained by considering the cross-correlation between the reflection \(\omega_r^k=2(\omega_i^k\cdot n) n - \omega_i^k\) and the observation:
\begin{equation}
\begin{gathered}
\label{eqn:ns_cross_corr}
\hat{n}_s \!=\! \argmax_{\substack{n \\ n^\text{init}:=n_s^\text{init}}} \mathfrak{N}\left(\!\begin{bmatrix} \!\nu_s^0 \!\cdot\! \omega_r^0 \!\cdot\! \omega_o\! \\ \cdots \\ \!\nu_s^k \!\cdot\! \omega_r^k \!\cdot\! \omega_o\! \end{bmatrix}\!\right)^{\!T\!} \!\mathfrak{N}\left(\!\begin{bmatrix} \hat{I}_s^0 \\ \cdots \\ \hat{I}_s^k \end{bmatrix}\!\right) \\ \text{s.t.} \quad \lVert n \rVert=1 , n \cdot \omega_i^k > 0 \text{, and\,} \hat{I}_s^k \in \hat{\Lambda}_s
\end{gathered}
\end{equation}

In most cases, the surface normals obtained through cross-polarized OLAT and parallel-polarized OLAT tend to exhibit similarity. However, in cases where the material exhibits stronger energy absorption, often indicated by a darker appearance, the diffuse reflection weakens, leading to inaccuracies in the diffuse normal. Also, when dealing with materials of a more intricate structure, inter-reflections and self-occlusion occur more frequently during the capture process, resulting in inaccurate specular normals. 

The optimization in Equations \ref{eqn:nd_cross_corr} and \ref{eqn:ns_cross_corr} assesses the alignment between normals and observations, allowing us to further improve normal quality. These cross-correlation coefficients, forming vectors in $\mathbb{R}^2$ for each pixel, are normalized and serve as blending weights for enhancing the measured normals, as shown in Fig.~\ref{fig:normal_fusion}.

\begin{figure}[t]
    \centering
    \vspace{-20pt}
    \includegraphics[width=\linewidth]{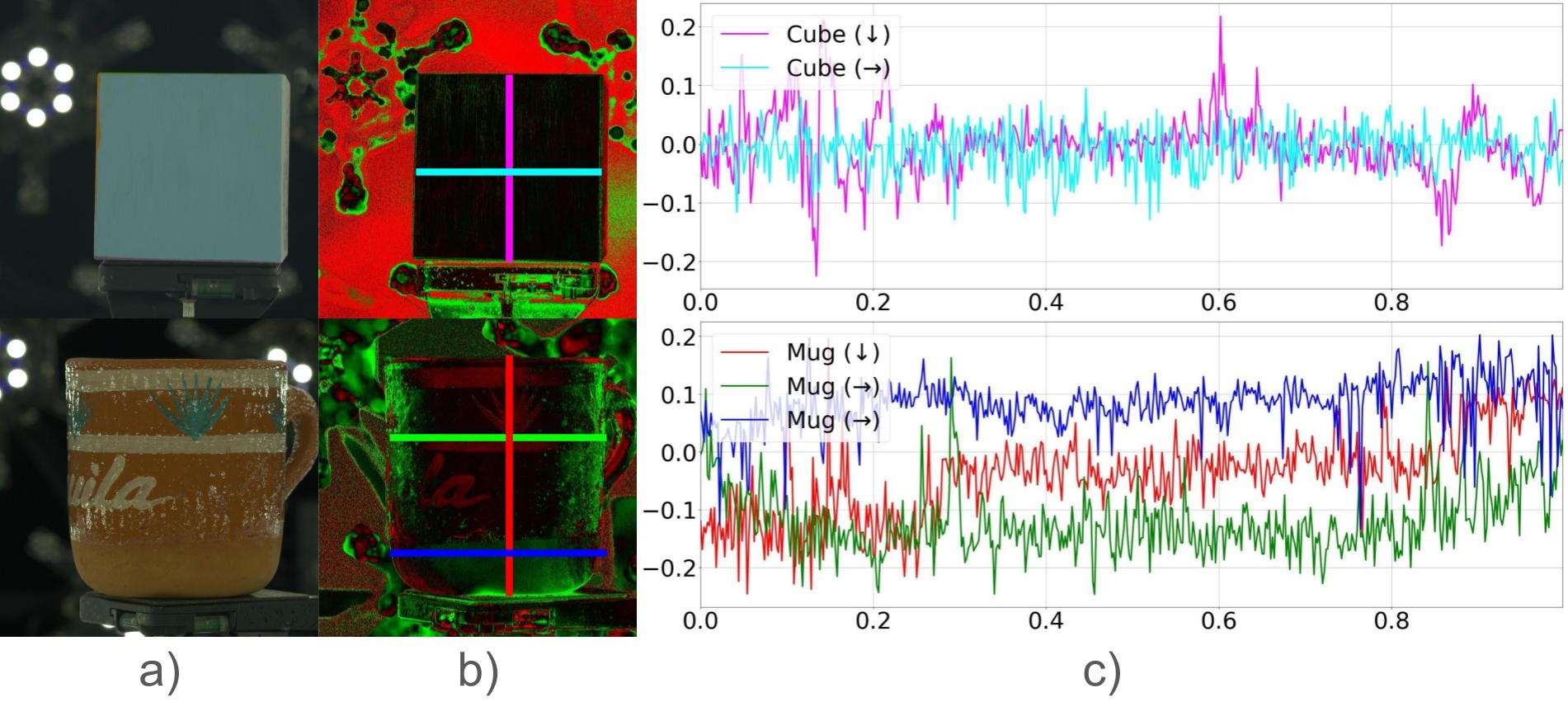}
    \vspace{-20pt}
    \caption{\textbf{Anisotropy Analysis.} We showcase a) the original capture, b) the anisotropy map $\mathbf{\varsigma}$, and c) the anisotropy variation along both vertical and horizontal directions as indicated by colored arrows on two objects. In the chart, the horizontal axis represents the pixel index, while the vertical axis is the measured anisotropy.}
    \vspace{5pt}
    \label{fig:anisotropy_analysis}
\end{figure}

\vspace{-30pt}
\paragraph{Anisotropy $\varsigma$ and Roughness $\gamma$}
Via captured OLAT, calculating the shape of the reflection lobe becomes straightforward, enabling the measurement of material isotropy/anisotropy. Ideally, the specular lobe follows a normal distribution governed by \(\sigma=(\sigma_x, \sigma_y)\), defined as \cite{ward1992measuring}:
\[
f_{\sigma}(\omega_i, \omega_o, n) \!=\! \frac{1}{4\pi\sigma_x\sigma_y\sqrt{ (\omega_o\!\cdot\!n)  (\omega_i\!\cdot\!n)}}\exp\!{\left(\!-2\frac{\frac{h\cdot t}{\sigma_x}^{\!2\!} \!+\! \frac{h\cdot b}{\sigma_y}^2}{1 \!+\! h\!\cdot\!n}\right)}
\]
where \(h=\mathfrak{N}(\omega_i+\omega_o)\) is the halfway vector, \([n, b, t]\) defines the local shading frame that aligns to the optimized specular surface normal $\hat{n}_s$. Following the collection of the response \(\hat{\Lambda}\), the optimization of the variance $\sigma$ of specular reflection, aimed at achieving the closest match to the observation, can be performed as:

\vspace{-10pt}
\begin{equation}
\label{eqn:sigma_optimization}
\begin{gathered}
\hat{\sigma} \!=\! \argmin_{\sigma \in \mathbb{R}_{+}^2} \!\left\lVert \mathfrak{N}\!\left(\!\begin{bmatrix} \!f_\sigma(\omega_i^0, \omega_o, \hat{n}_s)\! \\ \cdots \\ \!f_\sigma(\omega_i^k, \omega_o, \hat{n}_s)\! \end{bmatrix}\!\right)\! - \!\mathfrak{N}\!\left(\!\begin{bmatrix} \hat{I}_s^0 \\ \cdots \\ \hat{I}_s^k \end{bmatrix}\!\right) \!\right\rVert^2 \\
\text{s.t.} \quad \hat{n}_s \cdot \omega_i^k > 0 \text{, and\,}  \hat{I}_s^k \in \hat{\Lambda}_s
\end{gathered}
\end{equation}

When the material is isotropic, the specular lobe is symmetric, where \(\sigma_x \!\approx\! \sigma_y\). In such cases, diffuse material tends to have a flat and wide reflection lobe while that specular material is narrow and sharp. Material anisotropy $\varsigma$ and material roughness $\gamma$ can therefore be derived by:
\begin{equation}
\varsigma = \frac{\sigma_x-\sigma_y}{\sigma_x+\sigma_y},\quad \gamma = \lVert \sigma_x^2 + \sigma_y^2 \rVert
\end{equation}

\vspace{-15pt}
\paragraph{Refinement on $\rho$} The refined diffuse normal \(\hat{n}_d\) can be further employed to measure the diffuse albedo \(\hat{\rho}_d\) through:
\begin{equation}
\begin{gathered}
\hat{\rho}_d = \argmin_{\substack{\rho \\ \rho^\text{init}:=\rho_d^\text{init}}} \!\left\lVert \rho \!\cdot\! \begin{bmatrix} \hat{n}_d\!\cdot\!\omega_i^0 \\ \cdots \\ \hat{n}_d\!\cdot\!\omega_i^k \end{bmatrix}- \!\begin{bmatrix} \hat{I}_d^0 \\ \cdots \\ \hat{I}_d^k \end{bmatrix}\!\right\lVert^2 \\ 
\text{s.t.} \quad \hat{n}_d \cdot \omega_i^k > 0 \text{, and\,} \hat{I}_d^k \in \hat{\Lambda}_d
\end{gathered}
\end{equation}
Also, the specular albedo can be optimized similarly via the optimized specular normal $\hat{n}_s$, and optimized variance $\hat{\sigma}$:
\begin{equation}
\begin{gathered}
\hat{\rho}_s \!=\! \argmin_{\substack{\rho \\ \rho^\text{init}:=\rho_s^\text{init}}} \!\left\lVert \rho \!\cdot\! \begin{bmatrix} \!f_{\hat{\sigma}}(\omega_i^0, \omega_o, \hat{n}_s)\! \\ \cdots \\ \!f_{\hat{\sigma}}(\omega_i^k, \omega_o, \hat{n}_s)\! \end{bmatrix} - \!\begin{bmatrix} \hat{I}_s^0 \\ \cdots \\ \hat{I}_s^k \end{bmatrix}\!\right\rVert^2
\\ \text{s.t.} \quad \hat{n}_s \cdot \omega_i^k > 0 \text{, and\,} \hat{I}_s^k \in \hat{\Lambda}_s
\end{gathered}
\end{equation}
Additionally, the presence of shadows in the albedo can be partially mitigated by factoring in occlusion through $\hat{\rho}/\tau$, where \(\hat{\rho}\!\in\! \{\hat{\rho}_d, \hat{\rho}_s\}, \tau\!\in\! \{\tau_d, \tau_s\}\). However, it's important to acknowledge that $\hat{\rho}/\tau$ is vulnerable to noise.

\section{Results and Experiments}
\label{sec:experiments}
\begin{figure}[t]
    \centering
    \vspace{-22pt}
    \includegraphics[width=\linewidth]{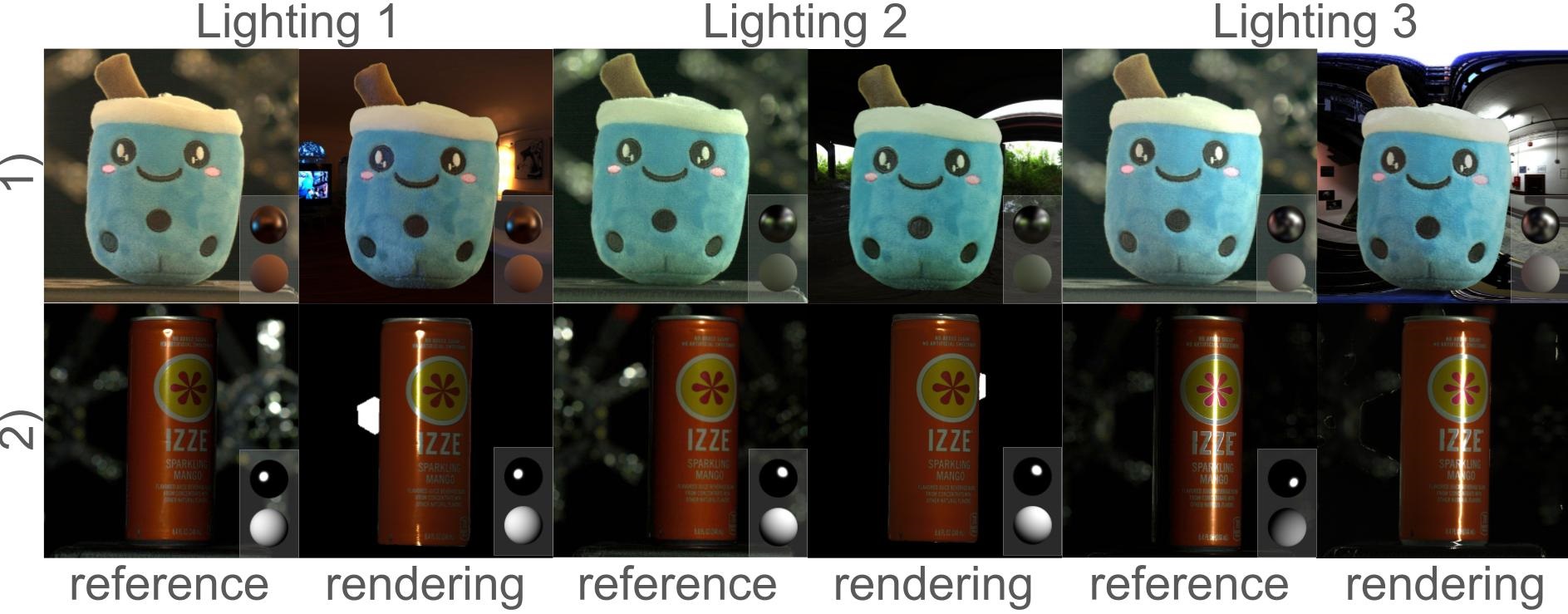}
    \vspace{-15pt}
    \caption{\textbf{Rendering Comparisons.} We validate our results using physically-based renderings with the measured materials. For each object, we showcase the renderings with corresponding captures under 1) environmental lighting or 2) area lighting conditions. 
    }
    \label{fig:supp_relighting}
    \vspace{5pt}
\end{figure}

\begin{figure*}[t]
    \centering
    \vspace{-20pt}
    \includegraphics[width=\linewidth]{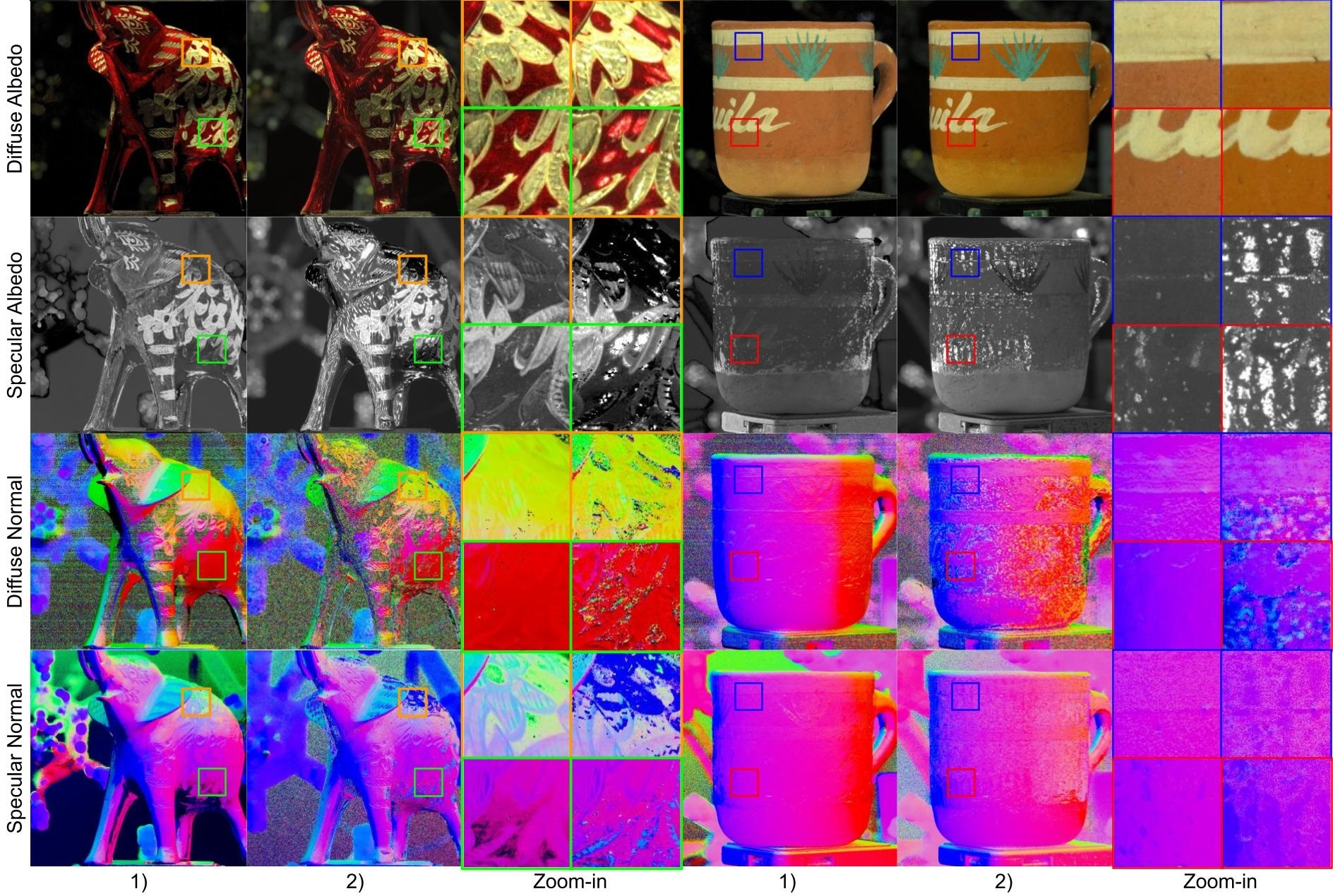}
    \vspace{-20pt}
    \caption{\textbf{Qualitative Comparison.} We compare results from 1) our method with 2) from \cite{ma2007rapid} via static capture on objects with specular outer layers. Examined properties cover diffuse albedo $\rho_d$ and specular albedo $\rho_s$, diffuse normal $n_d$, and specular normal $n_s$, with zoomed-in views. Normally, diffuse and specular normals are similar, but in multi-layered materials, they may differ slightly.}
    \vspace{-17pt}
    \label{fig:cmp_static_scan}
\end{figure*}

\begin{figure}[h]
    \centering
    \begin{minipage}{0.5\linewidth}
        \includegraphics[width=\linewidth]{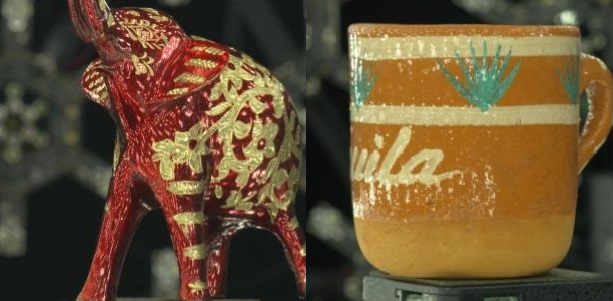}
    \end{minipage}%
    \hspace{5pt} 
    \begin{minipage}{0.4\linewidth}
        \caption{Reference Images for Figure \ref{fig:cmp_static_scan}. These are original captures with all-white lights on.}
        \label{fig:cmp_static_scan_reference}
    \end{minipage}
\end{figure}

\paragraph{Ablation Study} 
In Fig.~\ref{fig:ablation_overexposure} and Fig.~\ref{fig:ablation_optimization}, we highlight the importance of our proposed methods by showcasing the difference between utilizing or omitting the overexposure removal module and optimization. These steps are pivotal throughout the entire process: the removal of overexposure aids in cleanly separating diffuse and specular components, while optimized physically correct normals further enhance the quality of acquired material components afterward. 


\vspace{-5pt}
\paragraph{Anisotropy Distribution Analysis} We visualize the measured anisotropy values $\varsigma$ along specific directions in Fig.~\ref{fig:anisotropy_analysis}. For diffuse objects, the measured anisotropy across different directions is nearly zero, indicating that the reflection lobe is evenly distributed, demonstrating isotropic behavior. Conversely, for the specular object, the measured value tends to deviate, indicating anisotropy. In contrast, for specular objects, the measured values deviate, indicating anisotropy. For the mug, the red scanline crosses both the clearcoat and diffuse base, clearly distinguishing the two materials. The anisotropy variation along the scanline shows distinct splits, confirming our method's correctness. Furthermore, the renderings in Fig. \ref{fig:supp_relighting} of metal soda also validate this. Without anisotropy, the specular reflections on the cylindrical object would result in a spherical specular highlight rather than a linear strip.


\vspace{-5pt}
\paragraph{Relighting}
In Fig.~\ref{fig:supp_relighting}, we showcase image-based rendering achieved in Blender, using our measured material properties and various HDRI as global illumination. The reference image for HDRI illumination is synthesized by weighting the captured OLAT images for each light, with these weights derived from averaging the pixel values within corresponding spherical areas of the HDRI light probe. In Blender, we use the principal BSDF material, incorporating the measured diffuse albedo as the base color, the diffuse normal as the object's normal, and the measured specular albedo to set the specular IOR level along with the measured tangent. The measured roughness is also applied. Under HDRI lighting, our renderings closely match the reference image, and under area lighting conditions, our method accurately captures specular reflections on highly glossy materials.



\vspace{-10pt}
\paragraph{Qualitative Comparisons} We compare our results with \cite{ma2007rapid}, using synthesized gradient illumination from polarized OLAT capture, and static gradient illumination capture. The latter, designed for human skin, incorporates lighting pattern adjustments and lowers the lighting intensity to avoid lens flare and unexpected highlights, as shown in Fig.~\ref{fig:cmp_static_scan} with references in Fig.~\ref{fig:cmp_static_scan_reference}. \textbf{Due to space limits, more results can be found in the supplementary.}

Ma et al.~\cite{ma2007rapid} struggle to eliminate overexposure on object surfaces. Moreover, the mixture of specular reflection with diffuse reflection can compromise the quality of albedo. This may further affect the accuracy of captured normals when albedo is introduced in Equation \ref{eqn:ma2007_normals}. In contrast, our method achieves a distinct separation between the diffuse albedo and the specular albedo, effectively reducing overexposures on both maps and leading to a more accurate reflectance measurement. Furthermore, our approach enhances the captured diffuse normals and specular components while mitigating inaccuracies arising from albedo maps, accurately preserving intricate geometric details.

\vspace{-5pt}
\section{Conclusion}
\vspace{-5pt}
In this work, we introduce the polarized reflectance field for precise material acquisition. Our results showcase a comprehensive enhancement across various material layers through preprocessing and optimization, in alignment with physical principles. Nevertheless, certain challenges persist. System inaccuracies arise from severe inter-reflections, involving the intricate distinction between direct and inter-reflected light. 
More limitations are detailed in the supplementary material.
These challenges and limitations could potentially be addressed with a neural network in the future.

{
    \small
    \bibliographystyle{ieeenat_fullname}
    \bibliography{main}
}

\newcommand{\opensupplement}{
    \setcounter{section}{0}
    \renewcommand\thesection{\Alph{section}}
}
\newcommand{\closesupplement}{
    \renewcommand\thesection{\arabic{section}}
}

\newpage
\opensupplement

\section{Proof}
\label{sec:proof}

In this section, we provide detailed explanations for the equations discussed in the Preliminary and Method sections of the main paper. 

\vspace{-10pt}
\paragraph{Separating Diffuse and Specular Reflection.}
The Fresnel equation reveals that specular reflection maintains the incident light's polarization state, while diffuse reflection remains unpolarized. 
When the polarizer and analyzer are set perpendicular to each other (cross polarization), the analyzer blocks the specular reflection, allowing only the diffuse reflection to be measured as $I_\perp$. Conversely, when the polarizer and analyzer are aligned in parallel (parallel polarization), measured as $I_\parallel$, the specular reflection remains observable. 


The state of polarization of light can be represented by a Stokes vector $S=[S_0,S_1,S_2,S_3]^T$ \cite{collett2005field}, where $S_0$ is the total light intensity, $S_1$ is the difference in intensity between horizontal and vertical linear polarization, $S_2$ is the difference between linear polarization at $\frac{\pi}{4}$ and $-\frac{\pi}{4}$ and $S_3$ is the difference between right-hand and left-hand circular polarization. The transformation of light's polarization states via a linear polarizer at an angle $\theta$ relative to a reference axis is defined by Mueller Matrics, M:
\begin{equation}
\begin{split}
    S'&=MS \\
    M&=\frac{1}{2}\!\begin{bmatrix}
        1 & \cos{2\theta} & \sin{2\theta} & 0\\
        \cos{2\theta} & \cos^2{2\theta} & \cos{2\theta}\sin{2\theta} & 0\\
        \sin{2\theta} & \cos{2\theta}\sin{2\theta} & \sin^2{2\theta} & 0\\
        0 & 0 & 0 & 0
    \end{bmatrix}
\end{split}
\end{equation}
When unpolarized light is incident on the polarizer, the input Stokes vector is $S=[1,0,0,0]^T$. As the light passes through a horizontally oriented polarizer, where $\theta=0$, the polarization state of light transforms from $S$ to $S'$ defined as:
\begin{equation}
\begin{split}
    S'&=\frac{1}{2}\!\begin{bmatrix}
        1 & \cos{0} & \sin{0} & 0\\
        \cos{0} & \cos^2{0} & \cos{0}\sin{0} & 0\\
        \sin{0} & \cos{0}\sin{0} & \sin^2{0} & 0\\
        0 & 0 & 0 & 0
    \end{bmatrix} \begin{bmatrix}
        1 \\ 0 \\ 0 \\ 0
    \end{bmatrix} \\
    &=\begin{bmatrix}
        1 & 1 & 0 & 0
    \end{bmatrix}^T
\end{split}
\end{equation}
Upon passing through the analyzer with an angle $\varphi$ to the reference axis, the polarization state of the light further transforms from $S'$ to $S''$ defined as:
\begin{equation}
\begin{split}
    S''&=\frac{1}{2}\!\begin{bmatrix}
        1 & \cos{2\varphi} & \sin{2\varphi} & 0\\
        \cos{2\varphi} & \cos^2{2\varphi} & \cos{2\varphi}\sin{2\varphi} & 0\\
        \sin{2\varphi} & \cos{2\varphi}\sin{2\varphi} & \sin^2{2\varphi} & 0\\
        0 & 0 & 0 & 0
    \end{bmatrix} \begin{bmatrix}
        1 \\ 1 \\ 0 \\ 0
    \end{bmatrix} \\
    &=\frac{1}{2}\begin{bmatrix}
        1 + \cos{2\varphi} \\ 
        \cos{2\varphi} + \cos^2{2\varphi} \\ 
        \sin{2\varphi} + \cos{2\varphi}\sin{2\varphi} \\
        0
    \end{bmatrix} 
    =\frac{1+\cos{2\varphi}}{2} \begin{bmatrix}
        1 \\ 
        \cos{2\varphi} \\ 
        \sin{2\varphi} \\
        0
    \end{bmatrix} \\
    &=\cos^2{\varphi} \begin{bmatrix}
        1 & \cos{2\varphi} &\sin{2\varphi} & 0
    \end{bmatrix}^T
\end{split}
\end{equation}
Since the first number in $S''$ represents the total intensity, $\cos^2{\varphi}$ is the current intensity after the polarise and analyzer, which is also recognized as Malus's Law (refere to Equation 1). Additionally, when adjusting the analyzer's axis parallel ($\varphi=0$) or perpendicular ($\varphi=\frac{\pi}{2}$) to the polarizer, the transmitted light can be simplified as follows:
\begin{equation}
    S''=
    \begin{cases}
        \begin{bmatrix}
            1 & 1 & 0 & 0
        \end{bmatrix}^T & \text{if $\varphi=0$}\\
        0 & \text{if $\varphi=\frac{\pi}{2}$ }\\
    \end{cases} 
\end{equation}
At $\varphi=\frac{\pi}{2}$, the analyzer completely blocks the light. This characteristic can be leveraged to eliminate specular reflection. Therefore, the diffuse reflection can be represented via $I_\perp$ while the specular reflection can be represented as the difference of two measurements represented via $I_\parallel-I_\perp$. 

Moreover, on average, half of the light becomes polarized when passing through the polarizers \cite{collett2005field}, the total intensity from the diffuse reflection $I_d$ and specular reflection $I_s$ from the original unpolarized light can be derived via:
\begin{equation}
    \begin{split}
        I_\perp &= \frac{I_d}{2\pi} \int_0^{2\pi} \cos^2{\theta} d\theta = I_d \cdot \frac{1}{2} \\ 
        I_\parallel - I_\perp &= \frac{I_s}{2\pi} \int_0^{2\pi} \cos^2{\theta} d\theta \\ 
        I_d &= 2I_\perp \\
        I_s &= 2I_\parallel - 2I_\perp
    \end{split}
\end{equation}

\paragraph{Initial Albedo Estimation.} An initial estimation of diffuse and specular albedo $\rho_d^\text{init}$, $\rho_s^\text{init}$ can be roughly derived from the separated diffuse and specular reflection $I_d$, $I_s$. This estimation relies on the rendering equation for light transport at each surface point that inherently considers albedo. 

In the presence of uniform white illumination, the observed radiant intensities for both diffuse reflection $I_d$ and specular reflection $I_s$ at any given surface point $p$ can be described as follows:
\begin{equation}
    I_{d,p} = \int_{\omega_i \in \Omega} \rho_{d,p} \cdot L(\omega_i) \cdot (n_p \cdot \omega_i) \,d\omega_i
\end{equation}
\begin{equation}
    I_{s,p} = \int_{\omega_i \in \Omega} \rho_{s,p} \cdot L(\omega_i) \cdot f_\sigma(\omega_i, \omega_o) \,d\omega_i
\end{equation}
, where $\omega_i$ and $\omega_o$ represent the incoming and outgoing lighting directions, respectively, with $L(\omega_i)$ denoting the incident radiance from the $\omega_i$, which remains uniform over the sphere. Additionally, $\rho_{d,p}$ and $\rho_{s,p}$ correspond to the diffuse and specular albedo at point $p$, while $n_p$ represents the surface point normal, and $f_\sigma$ stands for the specular reflection distribution function.

We can establish a local frame where the Y-axis aligns with the surface normal $n$. Therefore, we can rewrite the $\omega_i$ in spherical coordinates as \(\omega_i = [\sin(\theta) \cos(\phi), \sin(\theta) \sin(\phi), \cos(\theta)]\), with $\theta$ as the polar angle and $\phi$ as the azimuthal angle within the local frame. Moreover, under uniform lighting conditions, the incident radiance $L(\omega_i)$ over the spherical sphere is a constant $L_0$. This allows us to further derive $\rho_d$ and $\rho_s$ over the entire observation as follows:
\begin{equation}
    \begin{split}
        \rho_d &= I_d \cdot \left(\int_{\omega_i \in \Omega} L(\omega_i) \cdot (n \cdot \omega_i) \,d\omega_i \right)^{-1} \\
        &= I_d \cdot \left( L_0 \cdot \int_{\omega_i \in \Omega} (n \cdot \omega_i) \,d\omega_i\right)^{-1} \\
        &= I_d \cdot \left(L_0 \cdot \int_{0}^{2\pi} \int_{0}^{\pi/2} \cos(\theta) \sin(\theta) \,d\theta d\phi \right)^{-1} \\
        &= I_d \cdot \left(L_0 \cdot \int_0^{2\pi} \frac{1}{2} d\phi \,\right)^{-1} \\
        &= I_d \cdot (L_0 \cdot \pi)^{-1} \\ 
    \end{split}
\end{equation}
\begin{equation}
    \begin{split}
        \rho_s &= I_s \cdot \left(\int_{\omega_i \in \Omega} L(\omega_i) \cdot f_\sigma(\omega_i, \omega_o) \,d\omega_i \right)^{-1} \\
        &= I_s \cdot \left( L_0 \cdot \int_{\omega_i \in \Omega} f_\sigma(\omega_i, \omega_o) \,d\omega_i\right)^{-1} \\
        &= I_s \cdot \left(L_0 \cdot 1 \right)^{-1}
    \end{split}
\end{equation}
Also, the $\int_{\omega_i \in \Omega} (n \cdot \omega_i) \,d\omega_i$ can also be simulated in the local frame via the law of large numbers:
\begin{equation}
\label{eqn:nDotL}
    \begin{split}
        y&= (2\pi)^{-1} \cdot \int_{\omega_i \in \Omega} (n \cdot \omega_i) \,d\omega_i \\
        &= \lim_{m\rightarrow \infty} \frac{1}{m} \sum_{k=0}^m \max(n \cdot \omega_i^k, 0) \\
        &= 2^{-1} \\
    \end{split}
\end{equation}
By conducting uniform sampling of $\omega_i$ from a unit sphere, we simulate the results as shown in Table \ref{tab:nDoti}.

\begin{table}[t]
    \centering
    \begin{tabular}{|c|cccccc}
         m&  $10^3$&  $10^4$&  $10^5$&  $10^6$&  $10^7$& $10^8$\\
         \hline
         y&  0.519&  0.503&  0.499&  0.499&  0.500& 0.500\\
    \end{tabular}
    \caption{Law of Large Numbers over Equation \ref{eqn:nDotL}. $m$ is the number of samples over the unit sphere and $y$ is the results.}
    \label{tab:nDoti}
\end{table}

Moreover, given that Ward's model \cite{ward1992measuring} subject to 2D normal distribution, denoted as $f_\sigma(\omega_i, \omega_o) \sim \mathcal{N}(\mu, \sigma^2)$. It's expected that the overall integral of $f_\sigma(\omega_i, \omega_o)$ over $\Omega$ equals 1. Further details can be found in \cite{ghosh2009estimating}.

During the capture, measurements become discrete through OLAT, with each light covering a specific area denoted as $A_0$ over the entire spherical surface. Consequently, we approximate the results via the captured sequence $\Lambda=\{I_d^i\}_{i=0}^N$: 
\begin{equation}
    \begin{split}
        \rho_d^\text{init} &= I_d \cdot \left(\int_{\omega_i \in \Omega} L(\omega_i) \cdot (n \cdot \omega_i) \,d\omega_i \right)^{-1} \\
        &= I_d \cdot (L_0 \cdot \pi)^{-1} \\
        &= 2\pi\frac{\int_{\omega_i\in \Omega} R_d(\omega_i, \omega_o) d\omega_i}{2\pi} \cdot (L_0 \cdot \pi)^{-1}\\
        &= \lim_{m \rightarrow \infty} \left( 2\pi\frac{\sum_m I_d^i}{m\cdot2^{-1}\cdot A_0} \right) \cdot (L_0 \cdot \pi)^{-1}\\
        &\approx 2\pi\frac{\sum_{i=0}^N I_d^i}{N\cdot2^{-1}\cdot A_0} \cdot (L_0 \cdot \pi)^{-1}\\
        &= \left( \frac{4\pi}{N \cdot A_0} \sum_{i=0}^N I_d^i \right) \cdot (L_0 \cdot \pi)^{-1}  \\
        &= \frac{4}{N \cdot L_0 \cdot A_0} \sum_{i=0}^N I_d^i \\
    \end{split}
\end{equation}
\begin{equation}
    \begin{split}
        \rho_s^\text{init} &= I_s \cdot \left(\int_{\omega_i \in \Omega} L(\omega_i) \cdot f_\sigma(\omega_i, \omega_o) \,d\omega_i \right)^{-1} \\
        &= I_s \cdot \left(L_0 \cdot 1 \right)^{-1} \\
        &= 2\pi\frac{\int_{\omega_i\in \Omega} R_s(\omega_i, \omega_o) d\omega_i}{2\pi} \cdot (L_0 \cdot 1)^{-1}\\
        &= \lim_{m \rightarrow \infty} \left( 2\pi\frac{\sum_m I_s^i}{m\cdot2^{-1}\cdot A_0} \right) \cdot \left(L_0 \cdot 1 \right)^{-1}\\
        &\approx \left( \frac{4\pi}{N \cdot A_0} \sum_{i=0}^N I_s^i \right) \cdot (L_0 \cdot 1)^{-1}  \\
        &= \frac{4\pi}{N \cdot L_0 \cdot A_0} \sum_{i=0}^N I_s^i \\
    \end{split}
\end{equation}
where $R_d$  and $R_s$ denote the diffuse reflection function and the specular reflection function of the material, respectively. $N$ is the number of lighting directions. For further simplification, we introduce the constant \(\kappa = (L_0 \cdot A_0)^{-1}\). As a result, the equation can be expressed as:
\begin{equation}
    \rho_d^\text{init} = \frac{4 \kappa}{N} \sum\nolimits_{i=0}^N I_d^i
\end{equation}
\begin{equation}
    \rho_s^\text{init} = \frac{4\pi \kappa}{N} \sum\nolimits_{i=0}^N I_s^i
\end{equation}

\vspace{-15pt}
\paragraph{Occlusion $\tau$}
Occlusion describes the overall visibility $\nu$  of incident lighting $\omega_i$ from the upper hemisphere $\Omega$ when observed via $\omega_o$. This is mathematically expressed as \(\frac{1}{2\pi}\int_\Omega \nu(\omega_i, \omega_o) \cdot (\omega_i\cdot n) d\omega_i\). Notably, since the average of $\omega_i \cdot n$ is $\frac{1}{2}$ over the hemisphere, the occlusion remains at $\frac{1}{2}$ when incident lighting $\omega_i$ from any solid angle is visible via $\omega_o$. To get the normalized occlusion over the hemisphere, we factor out the average of $\omega_i \cdot n$, resulting in:
\begin{equation}
    \begin{split}
        \tau &= \frac{1}{2\pi}\int_\Omega \nu(\omega_i, \omega_o) \cdot (\omega_i\cdot n) d\omega_i \cdot \left(\frac{1}{2\pi}\int_{\Omega} (\omega_i \cdot n) d\omega_i \right)^{-1} \\
        &= \frac{1}{2\pi}\lim_{m \rightarrow \infty} \left( 2\pi\frac{\sum_m I^i}{m\cdot2^{-1}} \right) \cdot (\frac{1}{2})^{-1}\\
        &\approx \frac{4}{N}\sum_{k=0}^{N} \lceil I^k-\zeta \rceil \cdot max(\omega_i^k\cdot n,0)
    \end{split}
\end{equation}
where $\zeta$ is the average ambient noise map and $\lceil \cdot \rceil$ is the ceiling operator.

\section{Implementation Details and Limitations}

\begin{figure}[t]
    \centering
    \includegraphics[width=\linewidth]{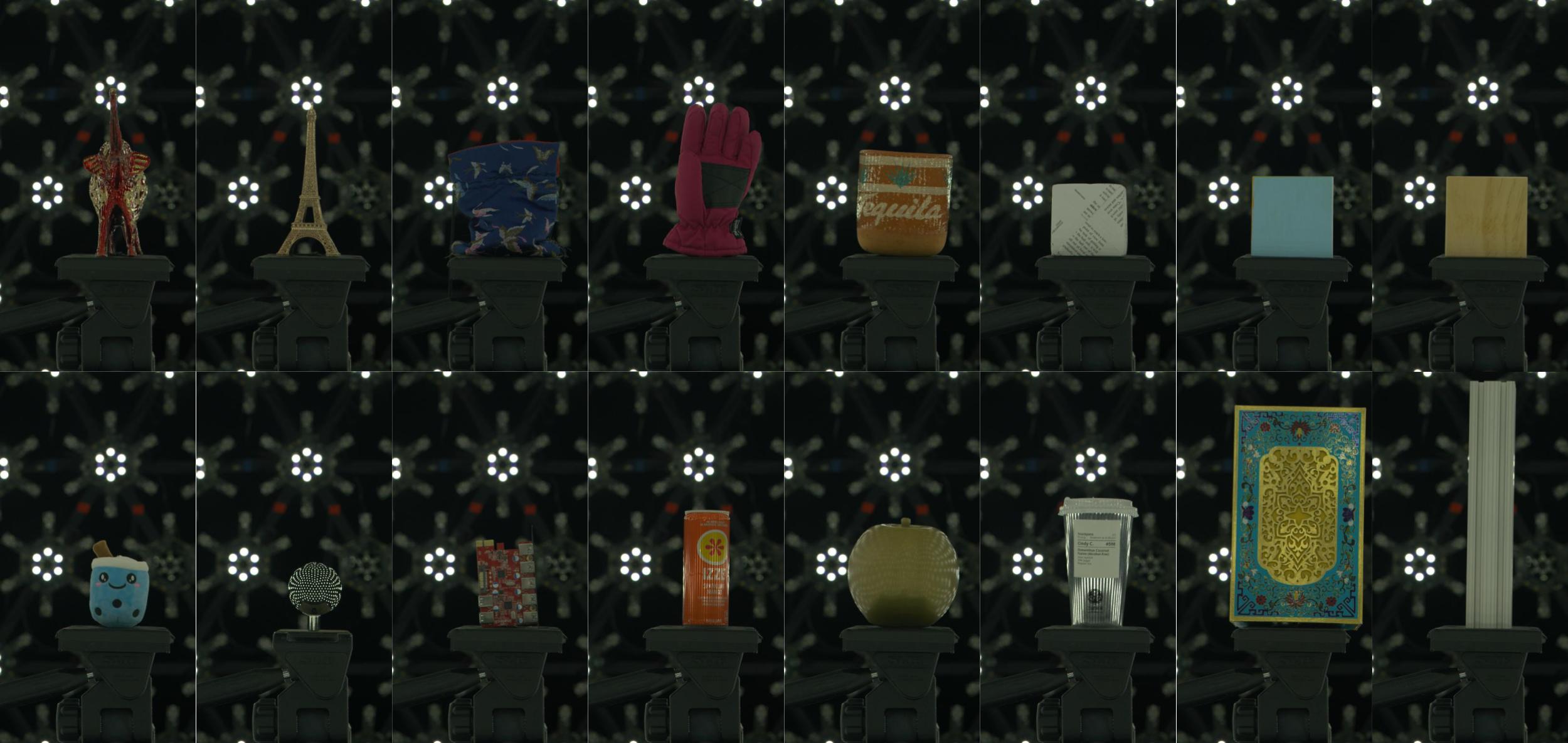}
    \caption{\textbf{Dataset Examples.}}
    \label{fig:supp_dataset_objects}
\end{figure}
\begin{figure*}[ht]
    \centering
    \includegraphics[width=\linewidth]{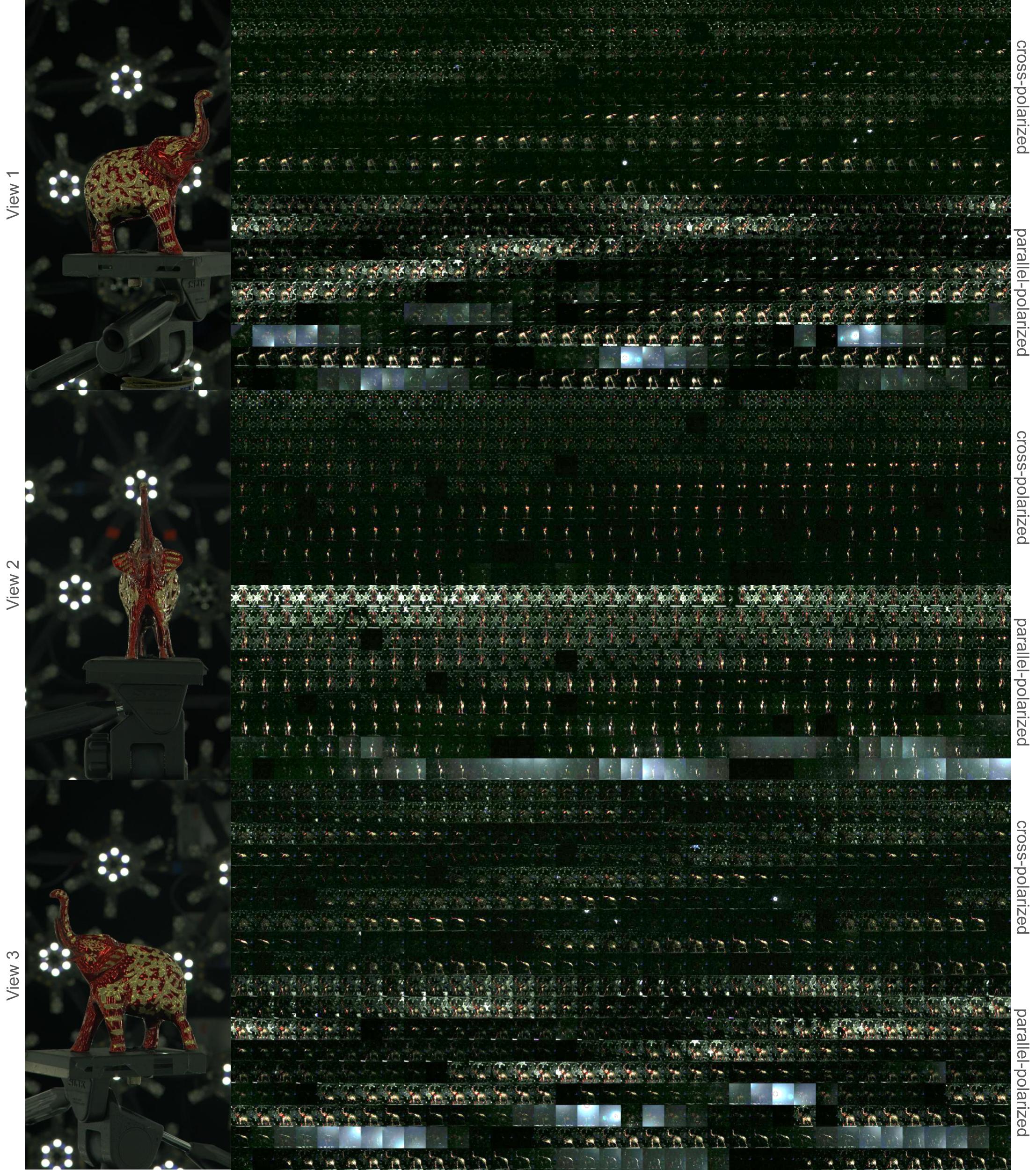}
    \caption{\textbf{Polarized OLAT from Multiviews.} We showcase an example captured object from multiview under cross-polarized and parallel-polarized OLAT.}
    \vspace{-10pt}
    \label{fig:supp_dataset_olat}
\end{figure*}

\paragraph{Storage and Memory Cost} Our capture process takes place in a Light Stage with 8 RED KOMODO 6K cameras, synchronized at 30 frames per second. For each scan, the captured data, saved as $3240 \times 6144$ resolution R3D files. Prior to using the data from the raw scan, we extract individual frames stored in OpenEXR format.  In this work, we conducted complete Polarized OLAT scans for 26 objects. We present frontal views of several objects in our dataset in Figure \ref{fig:supp_dataset_objects} and further present the polarized OLAT captures for a specific object from multiview in Figure \ref{fig:supp_dataset_olat}.


\paragraph{GPU Usage} Our implementation leverages JAX \cite{jax2018github} for efficient GPU access, ensuring a lightweight solution. We also employ JAXopt \cite{jaxopt_implicit_diff} for optimization, enabling batchable and differentiable solvers on large data blocks with a single workstation. We perform batch processing via vectorizing map on an RTX A6000, employing a batch size of $2^{20}$ for optimizing diffuse reflection and $2^{18}$ for optimizing specular reflection.

\vspace{-10pt}
\paragraph{Lens Flare and Inter-reflection Constrains} We implement the constraints $n \times \omega_i \geq 0$ to mitigate the impact of lens flare and inter-reflection. In practical terms, we apply these constraints to filter out values that do not meet the criteria before optimization. This approach reduces computational complexity and facilitates the optimizer's convergence to the optimal solution. Unconstrained data typically introduces a considerable number of zeros, which usually leads the optimizer to generate blank results. The constraints effectively address this issue.

\vspace{-10pt}
\paragraph{Noise Pixels} Throughout the optimization process, the presence of noise pixels (usually in the background) can significantly prolong the solver's search for the local minimum, and unfortunately, this extended search doesn't lead to a meaningful solution. To address this challenge, we choose to terminate the solver when the linear search encounters failure, which is often a consequence of noise pixels in the background.  Notably, the solver tends to converge more readily when dealing with pixels located in the foreground objects.

\paragraph{Limitations}
The proposed method successfully decomposes highly glossy materials. However, capturing living creatures or humans presents challenges due to the current setup's requirement to capture a dense reflectance field both with and without polarization. This process typically takes around 10 seconds given the current frame rate, and any slight movement of the subject can lead to color bleeding issues. Future implementations will need to incorporate frame tracking to address these movements effectively. Additionally, the method does not currently measure geometry, leading to difficulties in managing shadows caused by self-occlusion. This could be mitigated by integrating geometry data from multiview captures.
\section{More Experiments and Results}

\subsection{Ablation Studies}

\begin{figure*}[t]
    \centering
    \includegraphics[width=\linewidth]{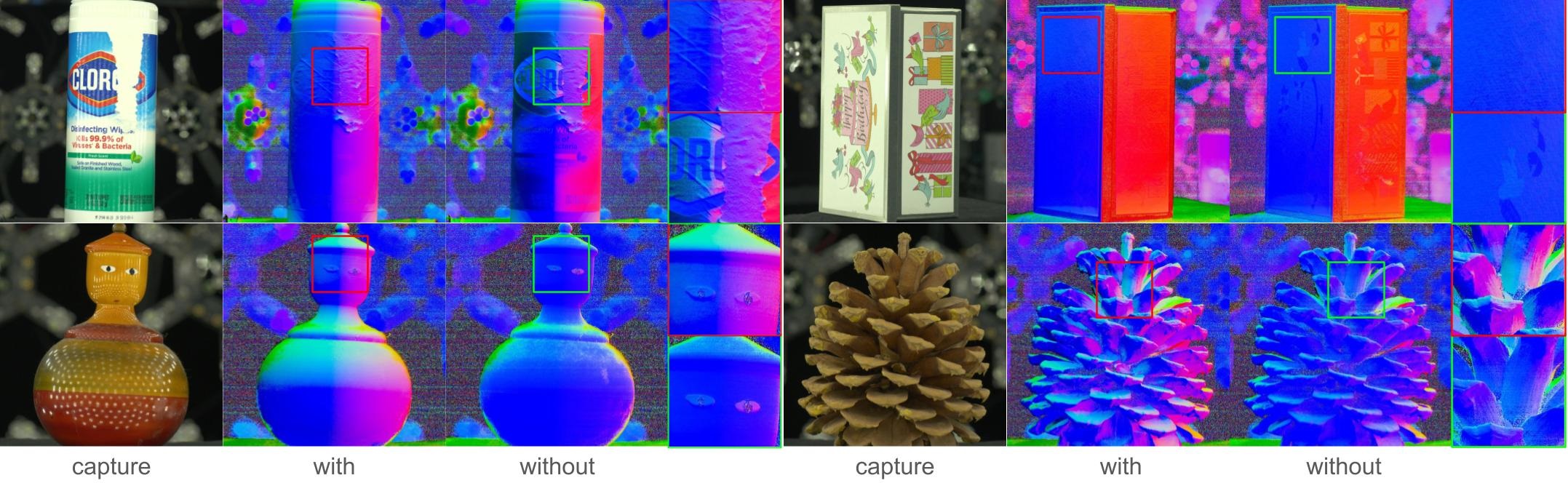}
    \vspace{-20pt}
    \caption{Ablation Study on acquired diffuse normal with and without optimization.}
    \label{fig:supp_ablation_optimization_diffuse_normals}
\end{figure*}

\begin{figure*}[t]
    \centering
    \includegraphics[width=\linewidth]{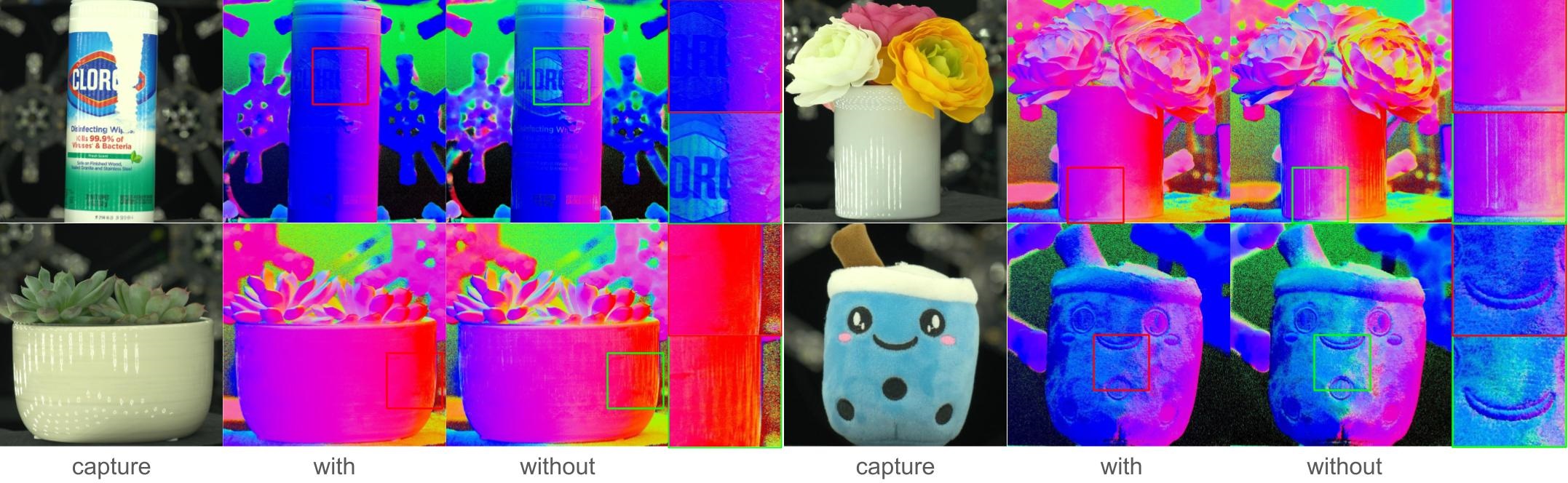}
    \vspace{-20pt}
    \caption{Ablation Study on acquired specular normal with and without optimization.}
    \label{fig:supp_ablation_optimization_specular_normals}
\end{figure*}

\begin{table}[t]
    \centering
    \small

    \setlength\tabcolsep{3pt} 
    
    \begin{tabular}{cccccc} 
          \hline
            Runtime $\downarrow$&$\hat{n}_d$&$\hat{\rho}_d$&$\hat{n}_s$&$\hat{\rho}_s$ &$\hat{\sigma}$\\
          \hline
  LBFGS (backtracking)& \textcolor{green}{1.54e-6}&1.63e-6&\textcolor{green}{1.75e-5}&8.21e-6&\textcolor{green}{2.06e-5}\\
  LBFGS (zoom)& 2.98e-6&3.37e-6&4.45e-5&1.43e-5&2.20e-5\\
  LBFGS (hager-zhang)& 7.94e-6&\textcolor{red}{4.81e-6}&5.79e-5&2.83e-5&\textcolor{red}{9.38e-5}\\
  GD& 3.01e-6& 2.23e-6&2.33e-5&\textcolor{green}{4.31e-6}&4.56e-6\\
  NCG& \textcolor{red}{3.39e-5}& 4.42e-6&\textcolor{red}{4.77e-4}&\textcolor{red}{6.75e-5}&6.78e-5\\
 GN& -& \textcolor{green}{8.33e-7}& -& 6.40e-6&5.97e-5\\
    \end{tabular}
    \begin{tabular}{cccccc} 
          \hline
            Error $\downarrow$&$\hat{n}_d$&$\hat{\rho}_d$&$\hat{n}_s$&$\hat{\rho}_s$ &$\hat{\sigma}$\\
          \hline
  LBFGS (backtracking) & 3.60e-4&\textcolor{red}{3.73}&2.02e-3&\textcolor{red}{3.94}&\textcolor{red}{7.31e-1}\\
  LBFGS (zoom)& 7.78e-5&5.31e-5&4.34e-4&7.87e-5&3.17e-1\\
  LBFGS (hager-zhang)& \textcolor{green}{7.64e-5}&4.85e-8&\textcolor{green}{3.29e-4}&2.49e-6&5.24e-2\\
  GD& \textcolor{red}{6.23e-2}& 5.12e-4&\textcolor{red}{2.54e-1}&5.13e-3&5.53e-1\\
  NCG& 2.20e-4& 2.74e-5&8.72e-4&1.13e-4&\textcolor{green}{1.28e-3}\\
 GN& -& \textcolor{green}{5.18e-10}& -& \textcolor{green}{1.16e-8}&2.21e-3\\
          \hline
    \end{tabular}
\vspace{-5pt}
\caption{\textbf{Solver Runtime (seconds) and Errors.} Lower is better ($\downarrow$). We emphasize the \textcolor{red}{max} and \textcolor{green}{min} in the column accordingly.}
\label{tab:solvers}
\end{table}
\paragraph{Ablation Study on Optimization} Further ablation results are presented in Figure \ref{fig:supp_ablation_optimization_diffuse_normals} and \ref{fig:supp_ablation_optimization_specular_normals} to affirm the quality enhancements achieved through optimization on both diffuse and specular normals. In the absence of optimization, the obtained normal maps may contain baked-in color artifacts, making the normal distribution sensitive to surface color variations. However, this issue is mitigated upon the introduction of optimization. Additionally, specular normals may exhibit blending with overexposure values, leading to noise in the data. The optimization process efficiently reduces such noise, contributing to overall improvement in quality.

\begin{figure*}[t]
    \centering
    \includegraphics[width=\linewidth]{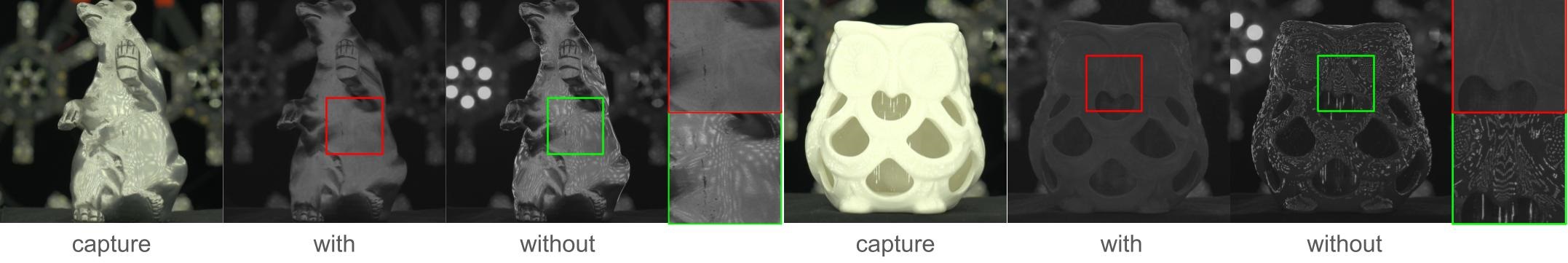}
    \vspace{-20pt}
    \caption{\textbf{Ablation study on overexposure removal: Specular Albedo}. The proposed overexposure removal effectively mitigates the lighting baked-in effect from the acquired specular albedo}
    \vspace{-10pt}
    \label{fig:supp_ablation_overexposure_specular_albedo}
\end{figure*}

\begin{figure*}[t]
    \centering
    \includegraphics[width=\linewidth]{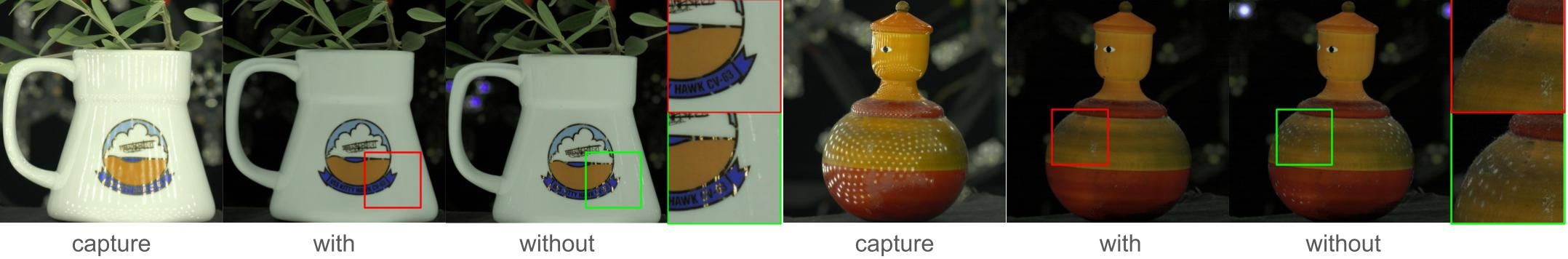}
    \vspace{-20pt}
    \caption{\textbf{Ablation Study on overexposure removal: Diffuse Albedo}. Similar to the improvements over specular albedo, overexposure removal can also improve diffuse albedo.}
    \vspace{-20pt}
    \label{fig:supp_ablation_overexposure_diffuse_albedo}
\end{figure*}

\begin{figure*}[t]
    \centering
    \includegraphics[width=\linewidth]{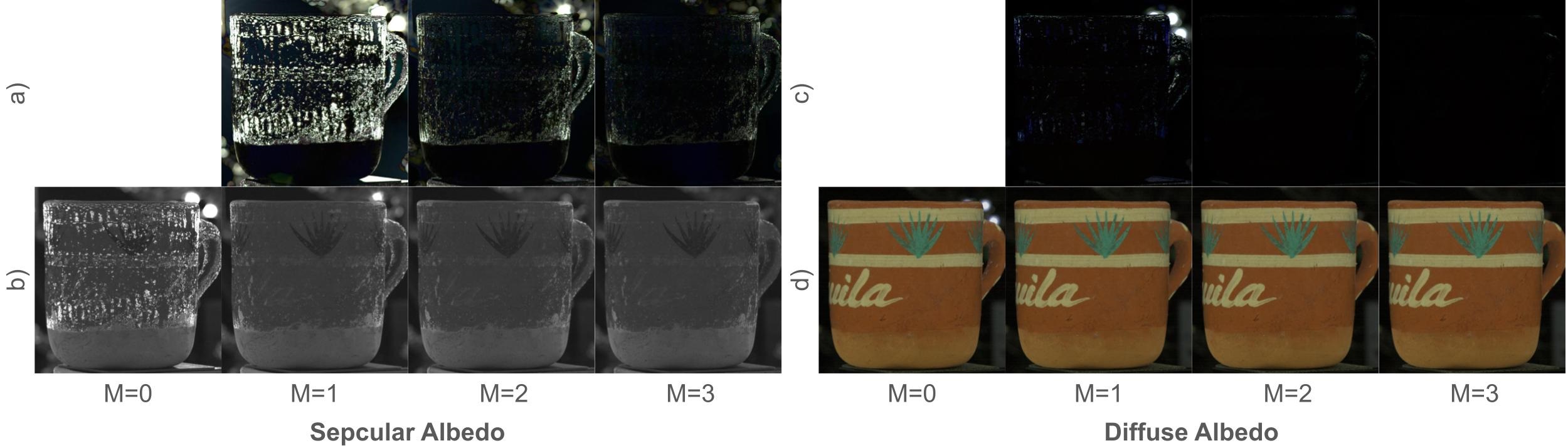}
    \vspace{-20pt}
    \caption{\textbf{Ablation Study on Overexposure Removal: Iterations} We showcase the diffuse albedo and specular albedo obtained with overexposure removal via various iterations in (b, d), and the corresponding removed overexposure values in (a, c), where M is the total number of iterations in the Overexposure Removal Algorithm.}
    \vspace{-20pt}
    \label{fig:supp_ablation_overexposure_step}
\end{figure*}

\vspace{-10pt}
\paragraph{Ablation Study on Overexposure Removal} Additional ablation results are presented in Figure \ref{fig:supp_ablation_overexposure_specular_albedo} and \ref{fig:supp_ablation_overexposure_diffuse_albedo}, illustrating the impact of overexposure removal on both specular albedo $\rho_s$ and diffuse albedo $\rho_d$. Particularly for objects with pronounced specular surfaces, our overexposure removal method effectively eliminates artifacts from incoming light sources while preserving intricate surface details. This efficacy is further demonstrated through visualizations showcasing different values of M, representing total iterations, in the Overexposure Removal Algorithm (see Figure \ref{fig:supp_ablation_overexposure_step}). 

The algorithm treats intensity variations as a sequential signal and addresses anomalies accordingly. Typically, during scanning, the maximum intensity values in the signal result from overexposure and offer limited useful reflection information. In practice, we set M=2 to efficiently remove overexposure while retaining the original intensity distribution to the maximum extent possible. This choice strikes a balance between removing overexposure artifacts and preserving valuable reflection data.

\vspace{-10pt}
\paragraph{Optimization Solvers} 
We compare the results obtained from various optimization solvers, including LBFGS (backtracking \cite{bazaraa2013nonlinear}), LBFGS (zoom \cite{nocedal1999numerical}), LBFGS (hager-zhang \cite{hager2005new}), Gradient Descent (GD), and Nonlinear Conjugate Gradient (NCG \cite{bazaraa2013nonlinear}), for solving $\hat{n}_d$, $\hat{\rho}_d$, $\hat{n}_s$, and $\hat{\rho}_s$. Additionally, we evaluate the performance of the Gauss-Newton (GN) nonlinear optimization approach for solving $\hat{\rho}_d$, $\hat{\rho}_s$, and $\hat{\sigma}$. In this context, optimizing surface normals $\hat{n}_d$ and $\hat{n}_s$ is inappropriate, as the cost function relies on correlation. The results are tested on RTX A6000 and averaged per pixel as shown in Table \ref{tab:solvers}. Note that errors are measured as the L2-norm of the gradient vector upon solver convergence or reaching the maximum iterations, 500 in our case. 

LBFGS (backtracking) achieves fast convergence but with higher errors, especially in albedo optimization, whereas LBFGS (zoom) and LBFGS (hager-zhang) require more time but offer improved accuracy. To balance runtime and error, we use LBFGS (backtracking) for normal optimization, LBFGS (zoom) for albedo optimization, and Gauss-Newton for $\sigma$ optimization. More implementation details can be found in the supplemental material.


\subsection{More Comparisons}

\begin{figure*}[h]
    \centering
    \includegraphics[width=\linewidth]{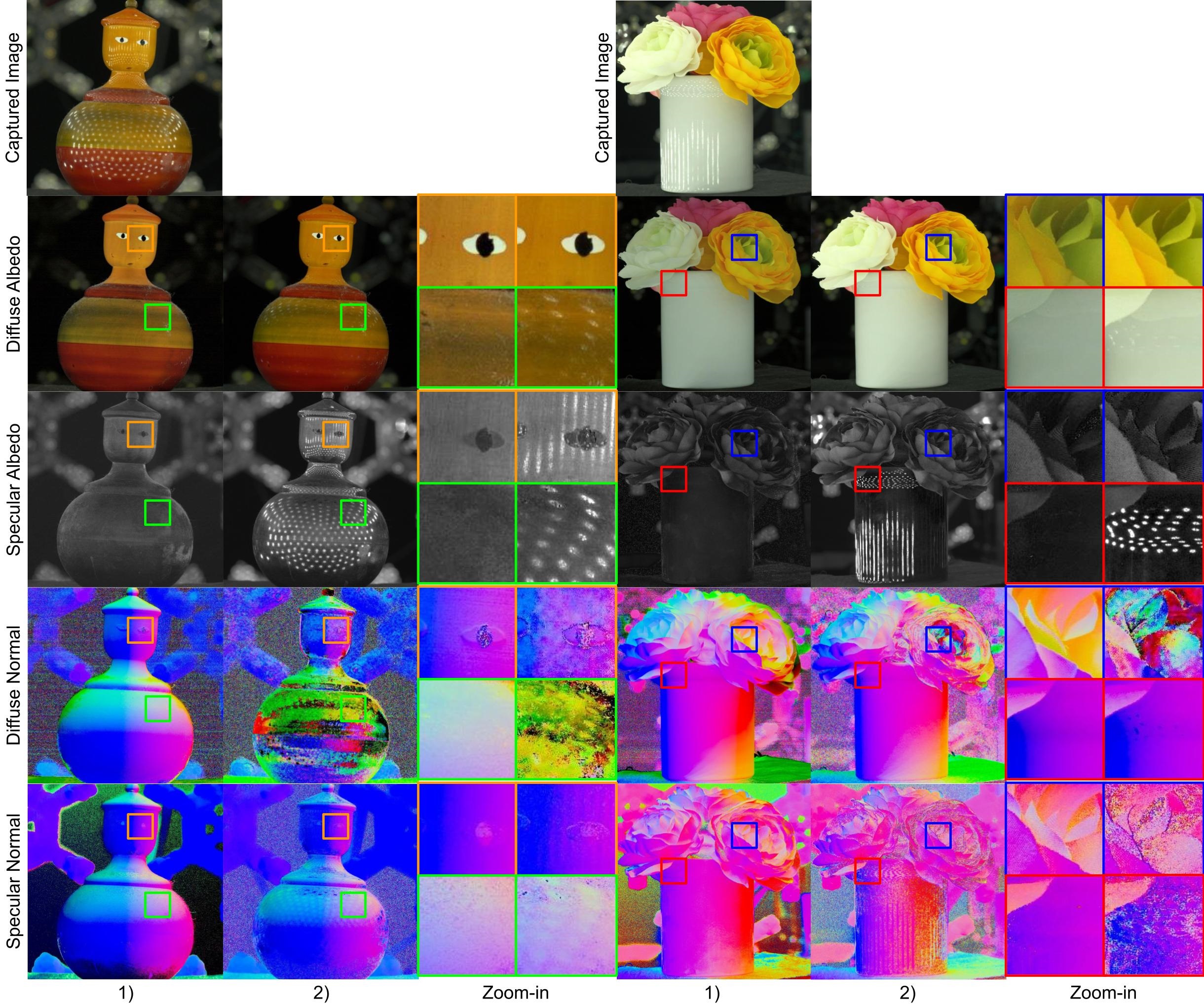}
    \caption{\textbf{More Qualitative Comparison.} We compare results from 1) our method with 2) from \cite{ma2007rapid} via static capture on objects with specular outer layers. Examined properties cover diffuse albedo $\rho_d$ and specular albedo $\rho_s$, diffuse normal $n_d$, and specular normal $n_s$, with zoomed-in views. Normally, diffuse and specular normals are similar, but in multi-layered materials, they may differ slightly. 
    }
    \vspace{-10pt}
    \label{fig:supp_cmp_static_scan_2}
\end{figure*}

\begin{figure*}[h]
    \centering
    \includegraphics[width=\linewidth]{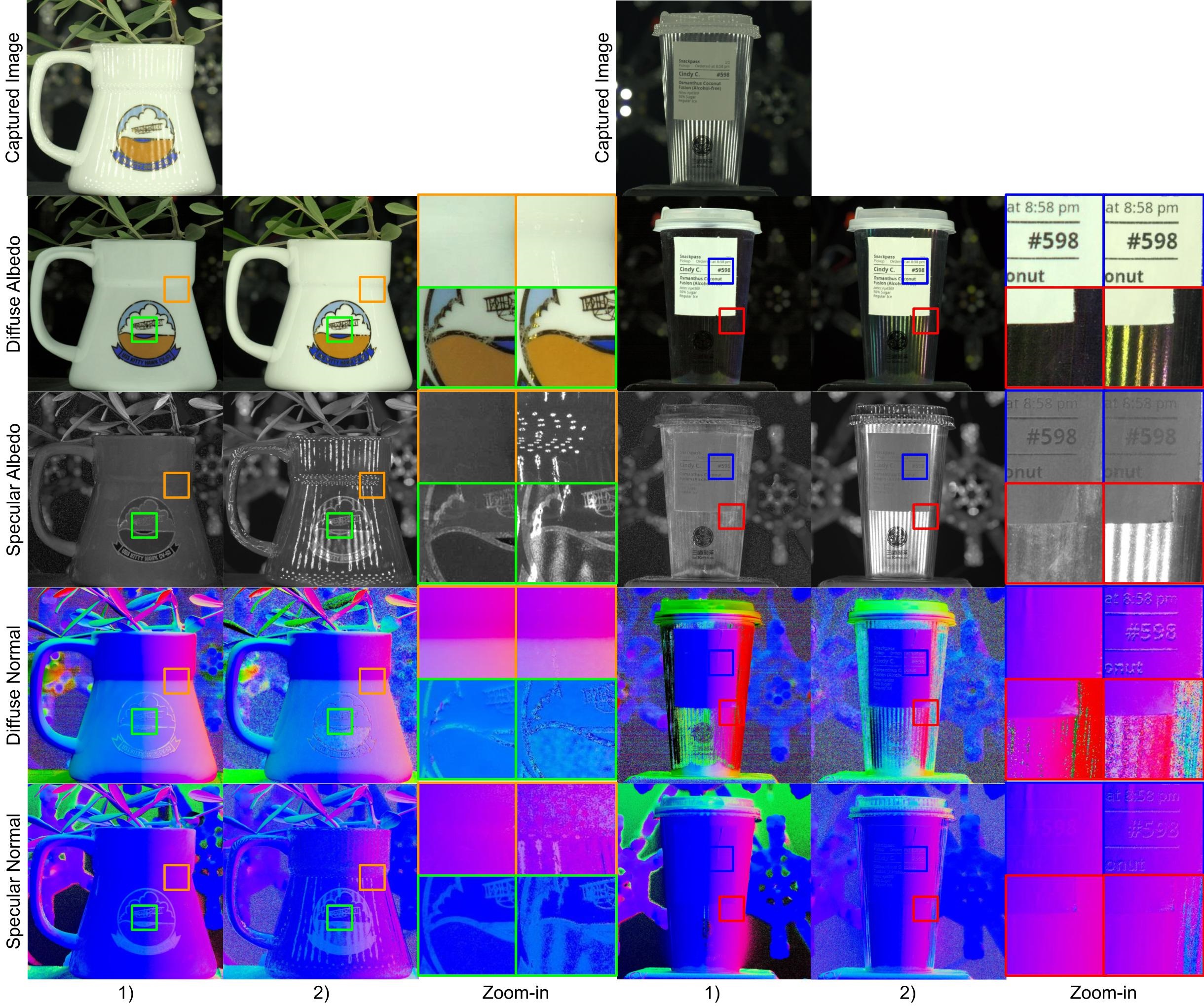}
    \caption{\textbf{More Qualitative Comparison.} We compare results from 1) our method with 2) from \cite{ma2007rapid} via static capture on objects with specular outer layers. Examined properties cover diffuse albedo $\rho_d$ and specular albedo $\rho_s$, diffuse normal $n_d$, and specular normal $n_s$, with zoomed-in views. Normally, diffuse and specular normals are similar, but in multi-layered materials, they may differ slightly. 
    }
    \vspace{-10pt}
    \label{fig:supp_cmp_static_scan}
\end{figure*}

In Figure \ref{fig:supp_cmp_static_scan_2} and \ref{fig:supp_cmp_static_scan}, we provide more qualitative comparisons between our results and \cite{ma2007rapid}, using static gradient illumination capture. 
Traditional methods encounter challenges when it comes to generating clean albedo for specular objects. Additionally, these methods struggle to effectively remove the undesired texture patterns from the measured normal maps. In contrast, our proposed method consistently surpasses traditional approaches, resulting in overall improved outcomes.

\subsection{Other Results}

\begin{figure*}[h]
    \centering
    \includegraphics[width=\linewidth]{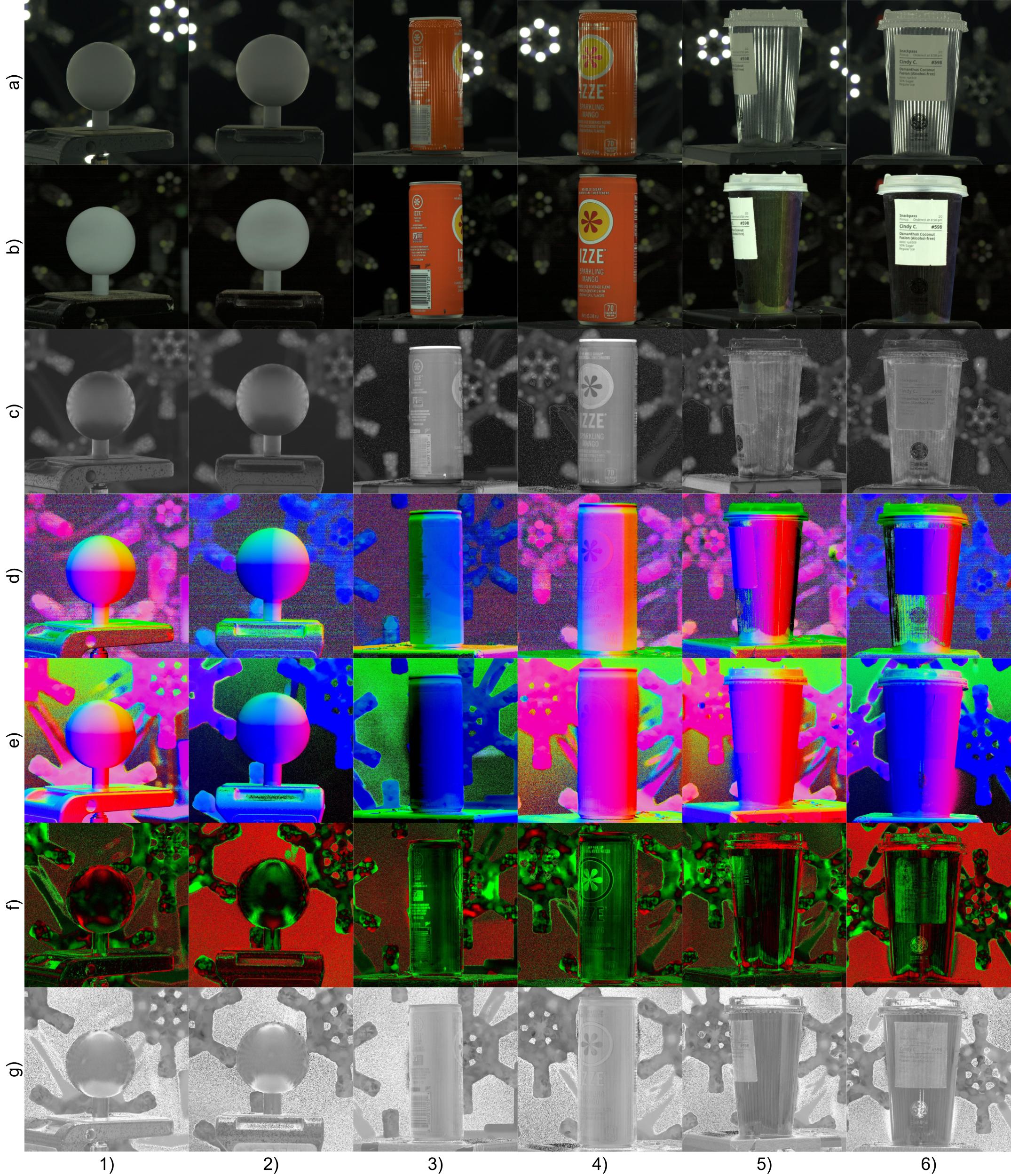}
    \caption{\textbf{More Optimization Results.} We present a grey ball from a) right and b) front, a soda can from c) left and d) right, and a specular cup from e) right and f) front. For each object, we showcase 1) original image, 2) diffuse albedo $\hat{\rho}_d$, 3) specular albedo $\hat{\rho}_s$, 4) diffuse normal $\hat{n}_d$, 5) specular normal $\hat{n}_s$, 6) anisotropy $\varrho$, and 7) roughness $\gamma$.}
    \label{fig:supp_qualitative_2}
\end{figure*}

\paragraph{More Optimization Results} As depicted in Figure \ref{fig:supp_qualitative_2}, this section provides additional results following our material optimization from multiple viewpoints. These results encompass a diverse range of shapes, from standard geometric forms to everyday objects, as well as materials spanning from relatively diffuse to highly specular. The materials demonstrate view consistency in the optimized maps and exhibit robustness against rotations.



\closesupplement

\end{document}